# ICRICS: Iterative Compensation Recovery for Image Compressive Sensing

Honggui Li[1*], Maria Trocan[2], Dimitri Galayko[3], Mohamad Sawan[4,5]
[1]School of Information Engineering, Yangzhou University, Yangzhou 225127, China
[2]LISITE Research Lab, Institut Supérieur d'Électronique de Paris, Paris 75006, France
[3]Laboratoire d'Informatique de Paris 6, Sorbonne University, Paris 75020, France
[4]Polystim Neurotech Laboratory, Polytechnique Montreal, Montreal H3T1J4, Canada
[5]School of Engineering, Westlake University, Hangzhou 310024, China
[1]hgli@yzu.edu.cn, [2]dimitri.galayko@sorbonne-universite.fr, [3]maria.trocan@isep.fr,
[4]mohamad.sawan@polymtl.ca, [5]sawan@westlake.edu.cn
[*]Corresponding Author

**Abstract**: Closed-loop architecture is widely utilized in automatic control systems and attain distinguished performance. However, classical compressive sensing systems employ open-loop architecture with separated sampling and reconstruction units. Therefore, a method of iterative compensation recovery for image compressive sensing (ICRICS) is proposed by introducing closed-loop framework into traditional compresses sensing systems. The proposed method depends on any existing approaches and upgrades their reconstruction performance by adding negative feedback structure. Theory analysis on negative feedback of compressive sensing systems is performed. An approximate mathematical proof of the effectiveness of the proposed method is also provided. Simulation experiments on more than 3 image datasets show that the proposed method is superior to 10 competition approaches in reconstruction performance. The maximum increment of average peak signal-to-noise ratio is 4.36 dB and the maximum increment of average structural similarity is 0.034 on one dataset. The proposed method based on negative feedback mechanism can efficiently correct the recovery error in the existing systems of image compressive sensing.

## 1 Introduction

Image compressive sensing simultaneously samples and compresses source signals, utilizes very low sub-Nyquist sampling frequency, depends on known image prior knowledge, and efficiently recovers original image. Image compressive sensing holds a series of prominent advantages, such as low sampling rate, low power consumption, low system cost, low radiation damage and high reconstruction performance, and achieves extensive applications in medical imaging, biological imaging, civil imaging and military imaging [1].

On a macro level, with the continuous development of human society and rapid progress of science and technology, people express higher and higher requirements on the performance of image compressive sensing systems, more and more disadvantages of image compressive sensing systems are exposed, the theories and applications of image compressive sensing obtain bigger development space, therefore it becomes the urgent need for scholars and engineers to research the principles and methods of high-performance image compressive sensing [2].

On a micro level, very low sampling rate results in the loss of vast useful information, and causes image compressed sensing to become an undetermined and ill-posed problem [3-5]. At present, this problem can only be approximately resolved and its perfect solution will possess very important value of theories and applications.

Currently, to the best of our knowledge, image compressive sensing adopts an open-loop architecture with separated measurement and recovery units. Actually, closed-loop architecture is broadly applied in automatic systems to gain better performance [6]. This paper attempts to enhance the recovery quality of the existing image compressive sensing approaches by building a closed-loop system. We name it as iterative compensation recovery for image compressive sensing (ICRICS). The main contributions of this papers are listed as follows: (1) a closed-loop negative feedback architecture is introduced into image compressive sensing to improve the reconstruction performance; (2) a theoretical analysis on the proposed negative feedback system is conducted; (3) an approximately mathematical proof of the effectiveness of the proposed method is provided; (4) the proposed method can be wielded for any existing approaches and strengthen their reestablishment capability.

The rest of this paper is organized as follows. Section 2 summarizes the related works of image compressive sensing, section 3 describes the theoretical foundation of the proposed method, section 4 designs the simulation experiments, and section 5 draws some conclusions.

## 2 Related-Work

Firstly, the theories and methods of image compressive sensing develop maturely, are still in the procedure of continuous improvement, and their application domains include magnetic resonance imaging, radar imaging, terahertz imaging, optical imaging, optoacoustic imaging, ghost imaging, ultrasound imaging, spectrometry imaging, hyperspectral imaging, microscopy imaging, underwater imaging, single-pixel imaging, single-photon imaging, Marchenko imaging, and etc. [7-27]. Li LX et al. summarize the sensing models, reconstruction algorithms and practical applications of compressive sensing [7]. Monika R et al. sum up the challenges, innovations and applications of adaptive block compressive sensing [8]. Chen QP et al. survey the technologies, applications and future prospects of compressive sensing for magnetic resonance imaging [9]. Bustin A et al. review the means of magnetic resonance imaging including low-rank reconstruction, sparse dictionary learning and deep learning [10]. Chul YJ et al. overview magnetic resonance imaging from the viewpoint of signal processing [11]. Yang JG et a. summarize the foundations, challenges and developments of compressive sensing for radar imaging [12]. Cao BH et al. probe the development history of terahertz imaging technologies [13]. Ke J et al. sum up the principles, advances, difficulties and opportunities of compressive sensing in optical imaging [14]. Hirsch L et al. compare time domain compressed sensing techniques for optoacoustic imaging [15]. Wang J et al. study the mathematical problems in ghost imaging [16]. Yousufi M et al. survey the applications of sparse representation based compressive sensing in ultrasound imaging [17]. Xie YR et al strengthen the throughput of mass spectrometry imaging using joint compressed sensing and subspace modeling [18]. Oiknine Y et al. investigate the technology of hyperspectral imaging [19]. Calisesi G et al. research microscopy imaging technology based on compressive sensing [20]. Monika R et al. present an efficient adaptive compressive sensing method of underwater imaging [21]. Edgar MP et al. discuss the principles and prospects of single-pixel imaging [22]. Xiao XY et al. review single-pixel imaging and its probability and statistics analysis [23]. Gibson GM et al. overview the recent developments, hardware configurations, mask designs, reconstruction algorithms and realistic applications of single-pixel imaging [24]. Zanotto L et al. survey the theories and technologies of single-pixel terahertz imaging [25]. Liu F et al. explore the technologies of single-photon imaging based on compressive sensing [26]. Zhang ML proposes compressive sensing acquisition for Marchenko imaging [27].

Secondly, image compressive sensing generally takes advantage of various prior knowledge to recover original image, and the classical methods comprise total variation, wavelet transform, sparse representation, low-rank representation, deep learning and etc. [28-58]. Ravishankar S et al. summarize the reconstruction methods of image compressive sensing from sparsity to data adaption method and machine learning [28]. Xie YT et al. review the deep learning methods and applications of compressed sensing image reconstruction [29]. Saideni W et al. overview the deep learning techniques for video compressive sensing [30]. Khosravy M et al. survey the random acquisition methods in compressive sensing [31]. Mishra I et al. review soft computing based compressive sensing techniques in signal processing [32]. Chen YT et al. systematically overview and analyze the fast method of magnetic resonance imaging based on artificial intelligence [33]. Zhang ML et al. utilize re-weighted total variation and sparse regression for magnetic resonance imaging relied on compressive sensing [34]. Zhang JC et al adopt 3D-total generalized variation and tensor decomposition for compressed sensing based magnetic resonance imaging [35]. Yin Z et al. propose multilevel wavelet-based hierarchical networks for image compressed sensing [36]. Yin Z et al. propose wavelet transform based deep network for image compressed sensing [37]. Lv MJ et al. employ low-rank representation for radar imaging [38]. Sun M et al. research the theory of magnetic resonance imaging based on blind compressive sensing and nonlocal low-rank constraint [39]. Li HG et al. apply sparse representation and compressive domain saliency based adaptive measurement to video compressive sensing [40]. Suantai S et al. study the applications of forward-backward algorithm in compressive sensing [41]. Shi WZ et al. propose convolutional neural network based image compressive sensing method including sampling network and reconstruction network where the former employs convolutional operation and the latter contains two parts: initial linear reconstruction and deep nonlinear reconstruction [2]. Yang Y et al. fuse traditional model based compressive sensing methods and data driven based deep learning methods [1]. Mardani M et al. present compressive sensing method of magnetic resonance imaging based on deep generative adversarial networks [42]. Li WZ et al. raise multi-scale generative adversarial network for image compressed sensing [43]. Zeng GS et al. review reconstruction technologies of magnetic resonance imaging based on non-full-sampling k-space deep learning [44]. Han Y et al. come up with acceleration technologies of magnetic resonance imaging based on low-rank k-space deep learning [45]. Kravets V et al. put forward progressive compressive sensing for large images based on deep learning [46]. Wang ZB et al. utilize multiscale deep network for compressive sensing image reconstruction [47]. Gan HP et al. employ data-driven acquisition and noniterative reconstruction for image compressive sensing [48]. Zhang J et al. present a range of deep learning based image compressive sensing methods including controllable arbitrary-sampling network (COAST), memory-augmented deep unfolding network (MADUN), interpretable optimization-inspired deep network (ISTA), ISTA+, ISTA++, optimization-inspired network (OPINE), and high-throughput deep unfolding network (HiTDUN) for image compressive sensing [49-54]. Zhang ZH et al. bring forward approximate message passing network (AMP-NET) for image compressive sensing [55]. The advanced transformer structure in deep learning has also been considered for image compressive sensing [56-58].

Thirdly, some methods, such as dictionary learning, super resolution, denoising, optimization and regularization, are utilized to upgrade the rebuilding capacity of image compressive sensing [59-63]. Harada Y et al. employ K-SVD dictionary learning to improve image quality for capsule endoscopy based on compressed Sensing [59]. Ueki W et al. adopt generative adversarial network-based image super resolution for accelerating brain magnetic resonance imaging [60]. Fang CJ et al. wield alternating direction method of multipliers for image denoising of compressed sensing [61]. El MA et al. utilize regularization constraints for image denoising of compressed sensing [62]. Pham CDK come up with enhancement method of compressive sensing image using multiple reconstructed signals [63]. Zhang Y. et al. propose a denoising method for compressed sensing based magnetic resonance imaging [64].

Finally, image compressive sensing seeks for the optimal balance between low sampling rate and high reestablishment capability, and its performance still has a big improvement space: sampling rate is expected to be further decreased and image recovery quality is expected to be further promoted. This paper tries to incorporate closed-loop negative feedback structure into image compressive sensing systems to further lift reconstruction performance.

## 3 Theory
### 3.1 Classical open-loop architecture

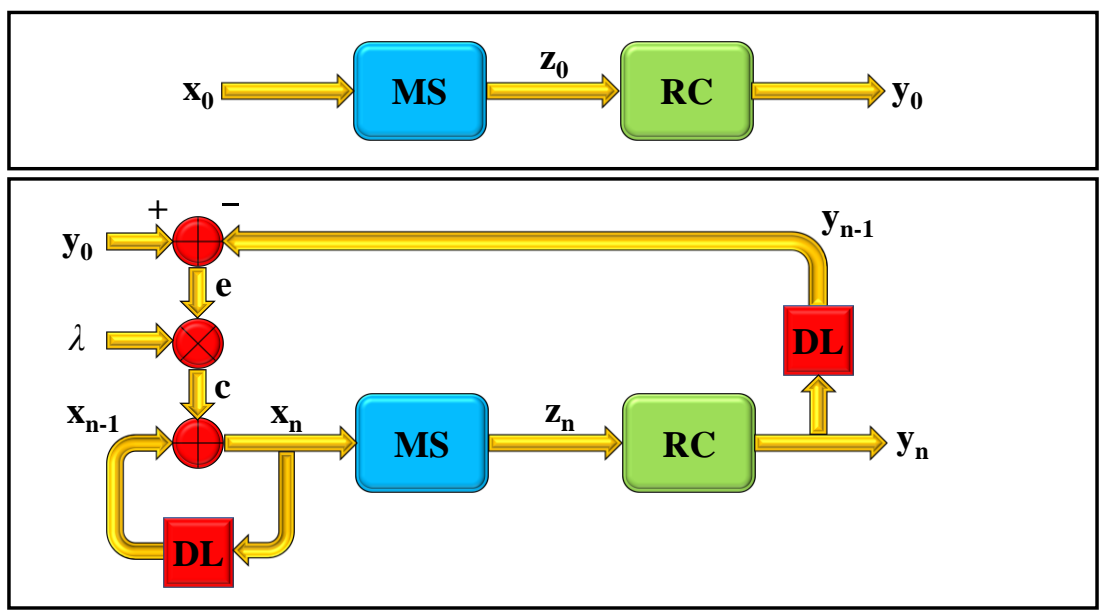

**Fig. 1 The classical and proposed architectures of image compressive sensing.**

The top of Fig. 1 is block diagram of classical open-loop image compressive sensing system which includes separated measurement (MS) unit and recovery (RC) units. MS samples original image signal and RC recovers it. The architecture can be described by following mathematical equations.

$$\begin{aligned} \mathbf{z}_0 &= \mathrm{MS}(\mathbf{x}_0) \\ \mathbf{y}_0 &= \mathrm{RC}(\mathbf{z}_0) \\ r &= \frac{d}{D} \\ d &< D \\ \mathbf{x}_0, \mathbf{y}_0 &\in R^D; \mathbf{z}_0 \in R^d \end{aligned} \quad (1)$$

where:
$\mathbf{x}_0$ is the input sample, i.e. the original image;
$\mathbf{z}_0$ is the measurement sample;
$\mathbf{y}_0$ is the output sample, i.e. the recovery image;
$r$ is the sub-Nyquist sampling rate;
$D$ is the dimension of original image;
$d$ is the dimension of measurement sample which is less than $D$.

The classical compressive sensing can be resolved by following optimization problem.

$$\begin{aligned} (\mathbf{z}_0, \mathbf{y}_0; \mathrm{MS}, \mathrm{RC}) &= \arg \min_{\mathbf{z}_0, \mathbf{y}_0; \mathrm{MS}, \mathrm{RC}} L(\mathbf{z}_0, \mathbf{y}_0; \mathrm{MS}, \mathrm{RC}) \\ \text{s.t.}\ \mathbf{z}_0 &= \mathrm{MS}(\mathbf{x}_0), \mathbf{y}_0 = \mathrm{RC}(\mathbf{z}_0) \end{aligned} \quad (2)$$

where $L$ is the loss function which is related to the prior knowledge of $\mathbf{z}_0$, $\mathbf{y}_0$, MS and RC, such as sparsity or deep learning based knowledge of $\mathbf{y}_0$.

### 3.2 Proposed closed-loop architecture

The bottom of Fig. 1 is schematic diagram of the proposed closed-loop ICRICS. It is based on any existing algorithms of image compressive sensing and takes the advantage of reconstruction sample $\mathbf{y}_0$ to restore the original image. It is composed of MS, RC, summator, multiplier and delayer (DL), and forms a closed-loop structure. It can be depicted by following iterative mathematical formulas.

$$\begin{aligned} \mathbf{x}_n &= \begin{cases} \mathbf{y}_0, n = 1 \\ \mathbf{x}_{n-1} + \mathbf{c}, n = 2, 3, \cdots\cdots \end{cases} \\ \mathbf{c} &= \mathbf{x}_{n-1} + \lambda \mathbf{e} \\ &= \mathbf{x}_{n-1} + \lambda(\mathbf{y}_0 - \mathbf{y}_{n-1}) \\ &= \mathbf{x}_{n-1} + \lambda(\mathbf{y}_0 - \mathrm{RC}(\mathbf{z}_{n-1})) \\ &= \mathbf{x}_{n-1} + \lambda(\mathbf{y}_0 - \mathrm{RC}(\mathrm{MS}(\mathbf{x}_{n-1}))) \end{aligned} \quad (3)$$

where:

$x_1$ is equal to $y_0$; can also be zero or a random value;

$x_n$ is expected to progressively approach $x_0$;

**c** is the control variable;

**e** is the error variable;

*n* is the number of iteration;

*λ* is a constant multiplicator.

The effectiveness of ICRICS in Eq. 2 can be approximately proven by following mathematical expressions. Actually, the correctness will be successfully verified in the experimental section below. Hereby, it is an attempt to establish an approximate mathematical proof.

$$
\begin{aligned}
\mathbf{x}_0 - \mathbf{x}_n &= \mathbf{x}_0 - (\mathbf{x}_{n-1} + \mathbf{c}) \\
&= \mathbf{x}_0 - (\mathbf{x}_{n-1} + \lambda \mathbf{e}) \\
&= \mathbf{x}_0 - (\mathbf{x}_{n-1} + \lambda (\mathbf{y}_0 - \mathbf{y}_{n-1})) \\
&= \mathbf{x}_0 - \mathbf{x}_{n-1} - \lambda (\mathbf{y}_0 - \mathbf{y}_{n-1}) \\
&= \mathbf{x}_0 - \mathbf{x}_{n-1} - \lambda (\mathrm{RC}(\mathbf{z}_0) - \mathrm{RC}(\mathbf{z}_{n-1})) \\
&= \mathbf{x}_0 - \mathbf{x}_{n-1} - \lambda (\mathrm{RC}(\mathrm{MS}(\mathbf{x}_0)) - \mathrm{RC}(\mathrm{MS}(\mathbf{x}_{n-1}))) \\
&\approx \mathbf{x}_0 - \mathbf{x}_{n-1} - \lambda \mathbf{G}(\mathbf{x}_0 - \mathbf{x}_{n-1}) \\
&= \mathbf{I}(\mathbf{x}_0 - \mathbf{x}_{n-1}) - \lambda \mathbf{G}(\mathbf{x}_0 - \mathbf{x}_{n-1}) \\
&= (\mathbf{I} - \lambda \mathbf{G})(\mathbf{x}_0 - \mathbf{x}_{n-1})
\end{aligned}
$$

$$\Rightarrow \|\mathbf{x}_0 - \mathbf{x}_n\|_2 \leq \|\mathbf{I} - \lambda \mathbf{G}\|_F \|\mathbf{x}_0 - \mathbf{x}_{n-1}\|_2$$

$$\mathbf{G} \to \mathbf{I}$$

$$\Rightarrow \|\mathbf{I} - \lambda \mathbf{G}\|_F \to \|\mathbf{I} - \lambda \mathbf{I}\|_F = \|\mathbf{I}(1-\lambda)\|_F = |1-\lambda|\|\mathbf{I}\|_F = |1-\lambda|\sqrt{D}$$

$$1 - \frac{1}{\sqrt{D}} < \lambda < 1 + \frac{1}{\sqrt{D}}$$

$$\Rightarrow |1-\lambda|\sqrt{D} < 1$$

$$\Rightarrow \|\mathbf{I} - \lambda \mathbf{G}\|_F \to |1-\lambda|\sqrt{D} < 1 \qquad (4)$$

$$\Rightarrow \|\mathbf{x}_0 - \mathbf{x}_n\|_2 < 1 \cdot \|\mathbf{x}_0 - \mathbf{x}_{n-1}\|_2 = \|\mathbf{x}_0 - \mathbf{x}_{n-1}\|_2$$

$$\Rightarrow \|\mathbf{x}_0 - \mathbf{x}_n\|_2 < \|\mathbf{x}_0 - \mathbf{x}_{n-1}\|_2 < \|\mathbf{x}_0 - \mathbf{x}_{n-2}\|_2 < \cdots\cdots < \|\mathbf{x}_0 - \mathbf{x}_1\|_2$$

$$\Rightarrow \|\mathbf{x}_0 - \mathbf{x}_n\|_2 \xrightarrow{n \to n_{\max}} 0$$

$$\Rightarrow \mathbf{x}_n \xrightarrow{n \to n_{\max}} \mathbf{x}_0$$

where:

RC(MS($x_{n-1}$)) is regarded as a nonlinear function, is expanded in Taylor series at $x_0$, and is approximatively linearized;

**I** is the unit matrix;

**G** is the matrix of Taylor series coefficients; it is an approximate diagonal matrix because it is reasonable to suppose that the pixels of an image are nearly independent and identically distributed;

$\|\cdot\|_F$ denotes Frobenus norm;

$\|\mathbf{I}-\lambda\mathbf{G}\|_F<1$ will be always met by assuming that **G** is an approximate diagonal matrix and choosing a suitable $\lambda$ belongs to numerical range $(1-D^{-1/2}, 1+D^{-1/2})$; $\lambda$ is equal to 1 in the latter experimental section for the sake of implementation convenience.

$n_{max}$ is the maximum number of iteration.

The bottom of Fig. 1 is a closed-loop global negative feedback system. The system input is $\mathbf{y}_0$ and the system output is $\mathbf{x}_n$, not $\mathbf{y}_n$. MS and RC constitute the controlled object. The top-left adder is the comparator which compares the difference between $\mathbf{y}_{n-1}$ and $\mathbf{y}_0$. The bottom-left adder accumulates the difference. If $\mathbf{y}_n$ increases firstly, **e** decreases, **c** decreases, $\mathbf{x}_n$ decreases and $\mathbf{y}_n$ decreases finally; and vice versa. After several cycles of comparison and adjustment, the system tends to stableness. According to the theory of negative feedback, $\mathbf{y}_n$ is close to $\mathbf{y}_0$ after the system is stable. Therefore, $\mathbf{x}_n$ should be also close to $\mathbf{x}_0$ after he system is stable to ensure that $\mathbf{y}_n$ is close to $\mathbf{y}_0$. It should be mentioned that the bottom-left delay unit and adder form a local positive feedback. It can be expressed by following mathematical formulas.

$$\begin{aligned} &\mathbf{y}_n \to \mathbf{y}_0 \\ &\mathbf{y}_0 = \mathrm{RC}(\mathrm{MS}(\mathbf{x}_0)) \\ &\mathbf{y}_n = \mathrm{RC}(\mathrm{MS}(\mathbf{x}_n)) \\ &\Rightarrow \mathbf{x}_n \to \mathbf{x}_0 \end{aligned} \tag{5}$$

If the proposed system can be approximately viewed as a linear system as aforementioned linear approximation in Taylor series, the related transfer functions can be depicted by following mathematical expressions.

$$\begin{aligned} \mathbf{MS}(s) &= \frac{\mathbf{Z}(s)}{\mathbf{X}_n(s)} \\ \mathbf{RC}(s) &= \frac{\mathbf{Y}_n(s)}{\mathbf{X}_n(s)} \\ \mathbf{DL}(s) &= \frac{\mathbf{X}_{n-1}(s)}{\mathbf{X}_n(s)} = \frac{\mathbf{Y}_{n-1}(s)}{\mathbf{Y}_n(s)} \\ \mathbf{H}_1(s) &= \frac{\mathbf{X}_n(s)}{\mathbf{C}(s)} = \frac{1}{\mathbf{I}-\mathbf{DL}(s)} \\ \mathbf{H}_g(s) &= \frac{\mathbf{Y}_n(s)}{\mathbf{Y}_0(s)} = \frac{\lambda\mathbf{H}_1(s)\mathbf{MS}(s)\mathbf{RC}(s)}{\mathbf{I}+\lambda\mathbf{H}_1(s)\mathbf{MS}(s)\mathbf{RC}(s)} \\ \mathbf{H}_e(s) &= \frac{\mathbf{E}(s)}{\mathbf{Y}_0(s)} = \frac{1}{\mathbf{I}+\lambda\mathbf{H}_1(s)\mathbf{MS}(s)\mathbf{RC}(s)} \\ \mathbf{e}_s &= \lim_{s\to 0} s\mathbf{IH}_e(s) \end{aligned} \tag{6}$$

where:

$s$ is the variable of complex frequency domain;

$\mathbf{X}_n$, $\mathbf{X}_{n-1}$, $\mathbf{Y}_n$, $\mathbf{Y}_0$ and $\mathbf{E}$ are the vectors of Laplace transforms of $\mathbf{x}_n$, $\mathbf{x}_{n-1}$, $\mathbf{y}_n$, $\mathbf{y}_0$ and $\mathbf{e}$;

**DL**, **MS** and **RC** are the matrices of transfer functions of DL, MS and RC;

$\mathbf{H}_1$ is the matrix of transfer function of local positive feedback;

$\mathbf{H}_g$ is the matrix of transfer function of global negative feedback;

$\mathbf{H}_e$ is the matrix of error transfer function;

$\mathbf{e}_s$ is the vector of stable error.

## 4 Experiments

### 4.1 Experimental conditions

The destination of experimental section is to compare the image reconstruction performance between the proposed method and the competition approaches. The experimental software platform is PyTorch framework of deep learning on 64-bt operating system. The experimental hardware platform is a laptop computer with 2.6 GHz dual-core processor and 8 GB main memory. The key parameters of the proposed method are collected in Tab. 1.

Tab. 1. The parameters of the proposed method.

| Name | Value | Meaning |
| --- | --- | --- |
| $\lambda$ | 1 | constant multiplicator |
| $n_{\max}$ | 5 | total number of iteration |

### 4.2 Competition Methods

10 state-of-the-art competition methods and related datasets utilized in the experiments are gathered in Tab. 1. The number at the end of each dataset name is the total number of images in the dataset. The proposed method depends on the open-source codes of the competition approaches. The parameters of deep neural networks in the proposed method are same as those of the competition approaches.

Tab. 2. Competition Methods.

| Abbreviations | Datasets | References |
| --- | --- | --- |
| COAST | Set5, Set11, Set14, BSD68 | [49] |
| MADUN | Set5, Set11, Set14 | [50] |
| ISTA | Set5, Set11, Set14 | [51] |
| ISTA+ | Set5, Set11, Set14, Brain50 | [51] |
| ISTA++ | Set5, Set11, Set14 | [52] |
| OPINE | Set5, Set11, Set14 | [53] |
| AMP-NET | Set5, Set11, Set14 | [55] |
| MTC-CSNET | Set5, Set11, Set14 | [56] |
| TCS-NET | Set5, Set11, Set14, McM18 | [57] |
| TransCS | Set5, Set11, Set14 | [58] |

### 4.3 Experimental Results

The experimental results are listed in Tab. 3 to 12. In each table, the average peak signal-to-noise ratio (PSNR) and average structural similarity (SSIM) of the proposed method and competition approaches on dataset Set5, Set11 and Set14 with different sampling rate are exhibited.

Firstly, it can be found that the proposed method outperforms the existing approaches in reconstruction performance almost in all datasets and sampling rates. Only in Tab. 3, 4, 5, 6 and 11, in several very low sampling rates, the proposed method cannot improve the recovery quality.

Secondly, it can be discovered that the reconstruction performance increases when the sampling rate increases. The reason is that the initial value of $x_n$, i.e. $y_0$, with high sampling rate is closer to $x_0$ than that with low sampling rate.

Thirdly, it can be seen that the reconstruction performance decreases when the image dataset size increases form dataset Set5 to Set14. This is due to the fact that large image dataset has large dynamic range of image pixels which are more difficult to rebuild.

Fourthly, it is indicated by Tab. 10 that the maximum average PSNR increment is 4.36 (=32.14-27.78) dB and the maximum average SSIM increment is 0.034 (=0.9180-0.8840) at the sampling rate of 0.25 on dataset Set14. That is to say, the proposed method based on MTC-CSNET approach holds the largest increment of PSNR and SSIM.

Finally, it manifested by Tab. 12 that the maximum average PSNR is 43.52 dB and the maximum average SSIM is 0.9853 with the sampling rate of 0.50 on dataset Set11. In other words, the proposed method based on TransCS approach possesses the best reconstruction performance.

In order to clearly show the difference between the proposed method and the competition approaches, average PSNR and average SSIM on dataset Set5, Set11 and Set14 with different sampling rates are respectively displayed in Fig. 2 to 7. The horizontal coordinate is sampling rate and the vertical coordinate is average PSNR or SSIM. The proposed method is based on the 10 competition approaches. It can be found that the proposed method holds the best reestablishment capability.

For the purpose of demonstrating the recovery quality of the proposed method based on MTC-CSNET approach which has the largest increment of PSNR and SSIM, the reconstruction images on dataset Set5, Set11 and Set14 at the sampling rate of 0.25 are respectively shown in Fig. 8, 9 and 10. In each figure, the left is the original image, the middle is the reconstruction image of MTC-CSNET approach, and the right is the reconstruction image of proposed method. Image subareas with significant differences are marked with red boxes. Only images with obvious-change subareas in the datasets are shown. It is reflected that the proposed method based on negative feedback mechanism can efficiently correct the recovery errors in the competition approaches.

For the sake of comparing the reconstruction performance between the proposed method and the 10 competition approaches, the reconstruction images of original images, parrots and monarch, on dataset Set11 with the sampling rate of 0.10 are displayed respectively in Fig. 10 and 11. The image area with big differences are also marked with red boxes. The proposed method in Fig. 10 is based on TransCS approach and the proposed method in Fig. 11 is based on OPINE approach. The numbers in the brackets are PSNR and SSIM. It is revealed that the proposed method holds the optimal reestablishment capability.

Tab. 3 Comparison with COAST [49].

| dataset | $R$ | Average PSNR | | Average SSIM | |
|---|---|---|---|---|---|
| | | original | proposed | original | proposed |
| Set5 | 0.10 | 30.50 | **30.56** | 0.8794 | 0.8794 |
| | 0.20 | 34.18 | **34.25** | 0.9298 | **0.9302** |
| | 0.30 | 36.38 | **36.59** | 0.9515 | **0.9523** |
| | 0.40 | 38.33 | **38.48** | 0.9645 | **0.9654** |
| | 0.50 | 40.21 | **40.42** | 0.9744 | **0.9754** |
| Set11 | 0.10 | 28.69 | **28.74** | 0.8618 | **0.8621** |
| | 0.20 | 32.54 | **32.59** | 0.9251 | **0.9257** |
| | 0.30 | 35.04 | **35.10** | 0.9501 | **0.9506** |
| | 0.40 | 37.13 | **37.23** | 0.9648 | **0.9654** |
| | 0.50 | 38.94 | **39.08** | 0.9744 | **0.9752** |
| Set14 | 0.10 | 27.41 | 27.41 | 0.7799 | 0.7799 |
| | 0.20 | 30.71 | **30.74** | 0.8672 | 0.8672 |
| | 0.30 | 33.10 | **33.20** | 0.9106 | **0.9110** |
| | 0.40 | 35.12 | **35.26** | 0.9369 | **0.9376** |
| | 0.50 | 36.94 | **37.13** | 0.9549 | **0.9557** |
| BSD68 | 0.10 | 26.28 | **26.30** | 0.7422 | 0.7422 |
| | 0.20 | 29.00 | **29.03** | 0.8413 | **0.8415** |
| | 0.30 | 31.06 | **31.10** | 0.8934 | **0.8938** |
| | 0.40 | 32.93 | **32.98** | 0.9267 | **0.9275** |
| | 0.50 | 34.74 | **34.82** | 0.9497 | **0.9505** |

Tab. 4. Comparison with MADUN [50].

| dataset | $r$ | Average PSNR | | Average SSIM | |
|---|---|---|---|---|---|
| | | original | proposed | original | proposed |
| Set5 | 0.10 | 26.13 | 26.13 | 0.7677 | 0.7677 |
| | 0.25 | 36.07 | **36.15** | 0.9478 | **0.9485** |
| | 0.30 | 37.34 | **37.41** | 0.9568 | **0.9573** |
| | 0.40 | 39.11 | **39.25** | 0.9680 | **0.9687** |
| | 0.50 | 40.64 | **40.95** | 0.9758 | **0.9769** |
| Set11 | 0.10 | 22.47 | **22.51** | 0.6968 | **0.6981** |
| | 0.25 | 34.80 | **34.87** | 0.9490 | **0.9495** |
| | 0.30 | 36.07 | **36.14** | 0.9582 | **0.9586** |
| | 0.40 | 37.85 | **37.96** | 0.9689 | **0.9693** |
| | 0.50 | 39.37 | **39.60** | 0.9763 | **0.9769** |
| Set14 | 0.10 | 23.27 | 23.27 | 0.6606 | 0.6606 |
| | 0.25 | 32.83 | **32.90** | 0.9028 | **0.9033** |

| | 0.30 | 34.11 | **34.18** | 0.9203 | **0.9207** |
| | 0.40 | 35.73 | **35.84** | 0.9411 | **0.9414** |
| | 0.50 | 37.17 | **37.38** | 0.9556 | **0.9562** |

**Tab. 5. Comparison with ISTA [51].**

| dataset | r | Average PSNR | | Average SSIM | |
|---|---|---|---|---|---|
| | | original | proposed | original | proposed |
| Set5 | 0.10 | 29.00 | 29.00 | 0.8351 | 0.8351 |
| | 0.25 | 34.00 | **34.32** | 0.9180 | **0.9259** |
| | 0.30 | 35.24 | **35.63** | 0.9368 | **0.9420** |
| | 0.40 | 37.34 | **37.60** | 0.9571 | **0.9573** |
| | 0.50 | 39.48 | **39.75** | 0.9703 | **0.9719** |
| Set11 | 0.10 | 26.26 | **26.36** | 0.7961 | **0.7991** |
| | 0.25 | 31.85 | **32.07** | 0.9161 | **0.9185** |
| | 0.30 | 33.09 | **33.35** | 0.9316 | **0.9345** |
| | 0.40 | 35.38 | **35.59** | 0.9533 | **0.9550** |
| | 0.50 | 37.42 | **37.73** | 0.9675 | **0.9692** |
| Set14 | 0.10 | 25.99 | 25.99 | 0.7270 | 0.7270 |
| | 0.25 | 30.29 | **30.38** | 0.8647 | **0.8668** |
| | 0.30 | 31.46 | **31.66** | 0.8900 | **0.8918** |
| | 0.40 | 33.60 | **33.80** | 0.9251 | **0.9259** |
| | 0.50 | 35.71 | **35.92** | 0.9480 | **0.9492** |

**Tab. 6. Comparison with ISTA+ [51].**

| dataset | r | Average PSNR | | Average SSIM | |
|---|---|---|---|---|---|
| | | original | proposed | original | proposed |
| Set5 | 0.10 | 29.07 | 29.07 | 0.8388 | 0.8388 |
| | 0.25 | 34.53 | **34.66** | 0.9306 | **0.9316** |
| | 0.30 | 35.79 | **35.96** | 0.9435 | **0.9450** |
| | 0.40 | 37.88 | **38.06** | 0.9607 | **0.9621** |
| | 0.50 | 39.83 | **40.09** | 0.9722 | **0.9737** |
| Set11 | 0.10 | 26.49 | 26.49 | 0.8036 | 0.8036 |
| | 0.25 | 32.44 | **32.50** | 0.9237 | **0.9247** |
| | 0.30 | 33.70 | **33.78** | 0.9382 | **0.9392** |
| | 0.40 | 36.02 | **36.14** | 0.9579 | **0.9589** |
| | 0.50 | 38.07 | **38.23** | 0.9706 | **0.9715** |
| Set14 | 0.10 | 26.13 | 26.13 | 0.7340 | 0.7340 |
| | 0.25 | 30.69 | **30.70** | 0.8737 | **0.8742** |
| | 0.30 | 31.82 | **31.86** | 0.8976 | **0.8981** |
| | 0.40 | 33.93 | **33.99** | 0.9293 | **0.9303** |
| | 0.50 | 35.98 | **36.13** | 0.9510 | **0.9521** |
| Brain50 | 0.10 | 34.63 | **34.71** | 0.9035 | 0.9035 |
| | 0.20 | 38.70 | **38.73** | 0.9484 | **0.9492** |

|  | 0.30 | 40.97 | **41.00** | 0.9639 | **0.9641** |
|  | 0.40 | 42.64 | **42.69** | 0.9729 | **0.9732** |
|  | 0.50 | 44.12 | **44.18** | 0.9792 | **0.9797** |

Tab. 7. Comparison with ISTA++ [52].

| dataset | r | Average PSNR | | Average SSIM | |
|---|---|---|---|---|---|
|  |  | original | proposed | original | proposed |
| Set5 | 0.10 | 30.02 | **30.28** | 0.8702 | **0.8716** |
|  | 0.20 | 33.94 | **34.01** | 0.9250 | **0.9259** |
|  | 0.30 | 36.27 | **36.41** | 0.9485 | **0.9498** |
|  | 0.40 | 38.13 | **38.37** | 0.9621 | **0.9637** |
|  | 0.50 | 39.95 | **40.34** | 0.9725 | **0.9744** |
| Set11 | 0.10 | 28.34 | **28.37** | 0.8530 | **0.8541** |
|  | 0.20 | 32.33 | **32.36** | 0.9216 | **0.9222** |
|  | 0.30 | 34.85 | **34.92** | 0.9477 | **0.9483** |
|  | 0.40 | 36.94 | **37.03** | 0.9627 | **0.9634** |
|  | 0.50 | 38.73 | **38.88** | 0.9727 | **0.9739** |
| Set14 | 0.10 | 27.32 | **27.33** | 0.7714 | **0.7724** |
|  | 0.20 | 30.72 | **30.75** | 0.8628 | **0.8635** |
|  | 0.30 | 33.07 | **33.16** | 0.9074 | **0.9080** |
|  | 0.40 | 34.98 | **35.13** | 0.9341 | **0.9350** |
|  | 0.50 | 36.78 | **36.96** | 0.9525 | **0.9536** |

Tab. 8. Comparison with OPINE [53].

| dataset | r | Average PSNR | | Average SSIM | |
|---|---|---|---|---|---|
|  |  | original | proposed | original | proposed |
| Set5 | 0.01 | 22.44 | **22.82** | 0.6145 | **0.6154** |
|  | 0.04 | 28.48 | **28.72** | 0.8396 | **0.8404** |
|  | 0.10 | 32.95 | **33.07** | 0.9202 | **0.9206** |
|  | 0.25 | 37.11 | **37.25** | 0.9589 | **0.9597** |
|  | 0.50 | 41.95 | **42.16** | 0.9818 | **0.9825** |
| Set11 | 0.01 | 20.02 | **20.47** | 0.5362 | **0.5399** |
|  | 0.04 | 25.52 | **25.74** | 0.7879 | **0.7904** |
|  | 0.10 | 29.81 | **29.93** | 0.8904 | **0.8922** |
|  | 0.25 | 34.81 | **34.97** | 0.9514 | **0.9527** |
|  | 0.50 | 40.19 | **40.46** | 0.9800 | **0.9811** |
| Set14 | 0.01 | 21.55 | **21.70** | 0.5397 | **0.5406** |
|  | 0.04 | 25.69 | **25.82** | 0.7217 | **0.7237** |
|  | 0.10 | 28.85 | **28.93** | 0.8347 | **0.8359** |
|  | 0.25 | 33.13 | **33.26** | 0.9228 | **0.9239** |
|  | 0.50 | 38.10 | **38.37** | 0.9678 | **0.9688** |

Tab. 9. Comparison with AMP-NET [55].

| dataset | r | Average PSNR | | Average SSIM | |
|---|---|---|---|---|---|
| | | original | proposed | original | proposed |
| Set5 | 0.10 | 32.44 | **32.54** | 0.9081 | **0.9090** |
| | 0.25 | 37.09 | **37.22** | 0.9557 | **0.9564** |
| | 0.30 | 38.18 | **38.35** | 0.9629 | **0.9639** |
| | 0.40 | 40.03 | **40.18** | 0.9733 | **0.9739** |
| | 0.50 | 41.87 | **42.02** | 0.9803 | **0.9810** |
| Set11 | 0.10 | 29.40 | **29.51** | 0.8779 | **0.8785** |
| | 0.25 | 34.63 | **34.76** | 0.9480 | **0.9487** |
| | 0.30 | 36.02 | **36.15** | 0.9586 | **0.9594** |
| | 0.40 | 38.27 | **38.41** | 0.9715 | **0.9721** |
| | 0.50 | 40.33 | **40.50** | 0.9804 | **0.9809** |
| Set14 | 0.10 | 28.89 | **28.98** | 0.8247 | **0.8258** |
| | 0.25 | 33.23 | **33.37** | 0.9181 | **0.9188** |
| | 0.30 | 34.39 | **34.54** | 0.9333 | **0.9342** |
| | 0.40 | 36.35 | **36.48** | 0.9535 | **0.9541** |
| | 0.50 | 38.29 | **38.44** | 0.9666 | **0.9672** |

**Tab. 10. Comparison with MTC-CSNET [56].**

| dataset | r | Average PSNR | | Average SSIM | |
|---|---|---|---|---|---|
| | | original | proposed | original | proposed |
| Set5 | 0.01 | 22.07 | **22.44** | 0.5903 | **0.5964** |
| | 0.04 | 26.33 | **27.11** | 0.7943 | **0.8030** |
| | 0.10 | 30.60 | **31.37** | 0.8983 | **0.9022** |
| | 0.25 | 33.95 | **35.76** | 0.9472 | **0.9504** |
| Set11 | 0.01 | 21.85 | **22.00** | 0.5718 | **0.5831** |
| | 0.04 | 25.57 | **26.01** | 0.7828 | **0.7936** |
| | 0.10 | 29.45 | **29.97** | 0.8932 | **0.8980** |
| | 0.25 | 33.73 | **34.69** | 0.9515 | **0.9543** |
| Set14 | 0.01 | 20.84 | **21.56** | 0.4927 | **0.5220** |
| | 0.04 | 22.93 | **25.01** | 0.6542 | **0.6948** |
| | 0.10 | 25.53 | **28.09** | 0.7893 | **0.8183** |
| | 0.25 | 27.78 | **32.14** | 0.8840 | **0.9180** |

**Tab. 11. Comparison with TCS-NET [57].**

| dataset | r | Average PSNR | | Average SSIM | |
|---|---|---|---|---|---|
| | | original | proposed | original | proposed |
| Set5 | 0.01 | 22.75 | 22.75 | 0.6001 | 0.6001 |
| | 0.04 | 27.55 | **27.63** | 0.8169 | **0.8197** |
| | 0.10 | 31.48 | **31.55** | 0.9064 | **0.9079** |
| | 0.25 | 35.85 | **35.90** | 0.9557 | **0.9563** |
| Set11 | 0.01 | 21.09 | 21.09 | 0.5505 | 0.5505 |
| | 0.04 | 25.46 | **25.50** | 0.7863 | **0.7881** |

| | 0.10 | 29.04 | **29.08** | 0.8834 | **0.8847** |
| | 0.25 | 33.94 | **33.99** | 0.9508 | **0.9514** |
| Set14 | 0.01 | 21.65 | **21.67** | 0.5219 | **0.5222** |
| | 0.04 | 25.26 | **25.33** | 0.7074 | **0.7095** |
| | 0.10 | 28.19 | **28.25** | 0.8283 | **0.8302** |
| | 0.25 | 32.24 | **32.31** | 0.9204 | **0.9213** |
| McM18 | 0.01 | 23.63 | **23.66** | 0.6144 | **0.6149** |
| | 0.04 | 27.54 | **27.60** | 0.7907 | **0.7926** |
| | 0.10 | 30.97 | **31.04** | 0.8913 | **0.8927** |
| | 0.25 | 35.89 | **35.96** | 0.9579 | **0.9584** |

**Tab. 12. Comparison with TransCS [58].**

| dataset | r | Average PSNR | | Average SSIM | |
| --- | --- | --- | --- | --- | --- |
| | | original | proposed | original | proposed |
| Set5 | 0.10 | 33.56 | **33.63** | 0.9244 | **0.9251** |
| | 0.25 | 38.26 | **38.28** | 0.9652 | **0.9654** |
| | 0.30 | 39.01 | **39.08** | 0.9693 | **0.9698** |
| | 0.40 | 41.40 | **41.47** | 0.9791 | **0.9794** |
| | 0.50 | 43.42 | **43.52** | 0.9850 | **0.9853** |
| Set11 | 0.10 | 29.54 | **29.62** | 0.8877 | **0.8880** |
| | 0.25 | 35.06 | **35.07** | 0.9548 | **0.9550** |
| | 0.30 | 35.62 | **35.68** | 0.9588 | **0.9595** |
| | 0.40 | 38.46 | **38.49** | 0.9737 | **0.9739** |
| | 0.50 | 40.49 | **40.52** | 0.9815 | **0.9816** |
| Set14 | 0.10 | 28.81 | **28.85** | 0.8343 | **0.8362** |
| | 0.25 | 33.38 | **33.48** | 0.9244 | **0.9250** |
| | 0.30 | 34.03 | **34.20** | 0.9349 | **0.9361** |
| | 0.40 | 36.69 | **36.84** | 0.9572 | **0.9577** |
| | 0.50 | 38.66 | **38.81** | 0.9693 | **0.9697** |

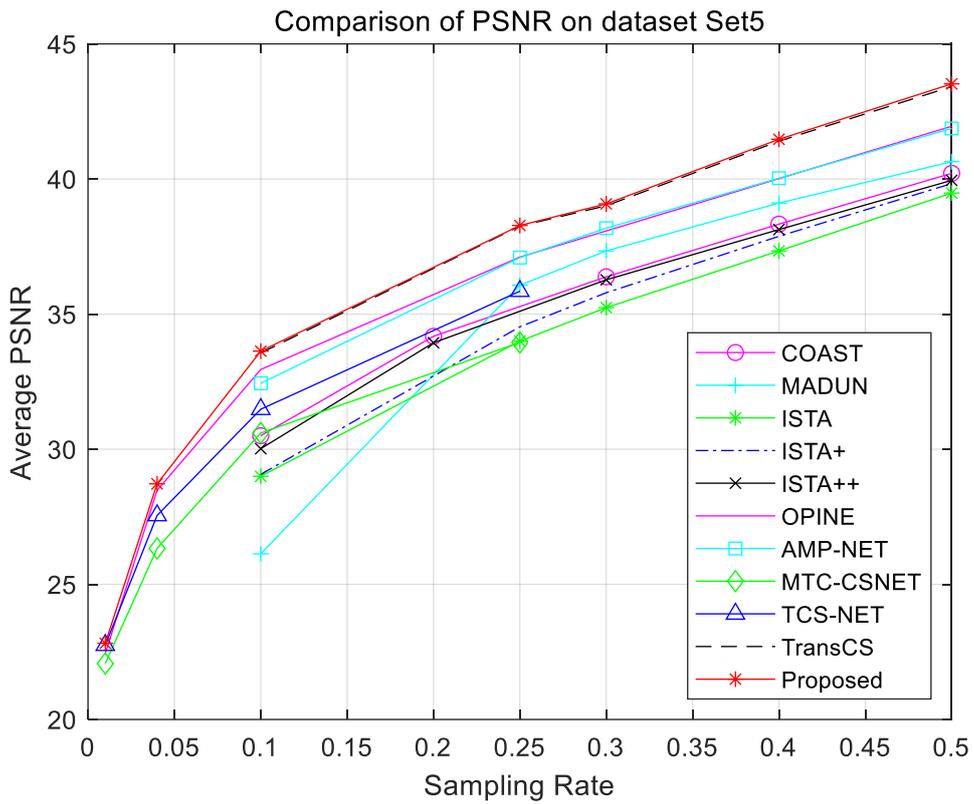

**Fig. 2.** Comparison of PSNR on dataset Set5.

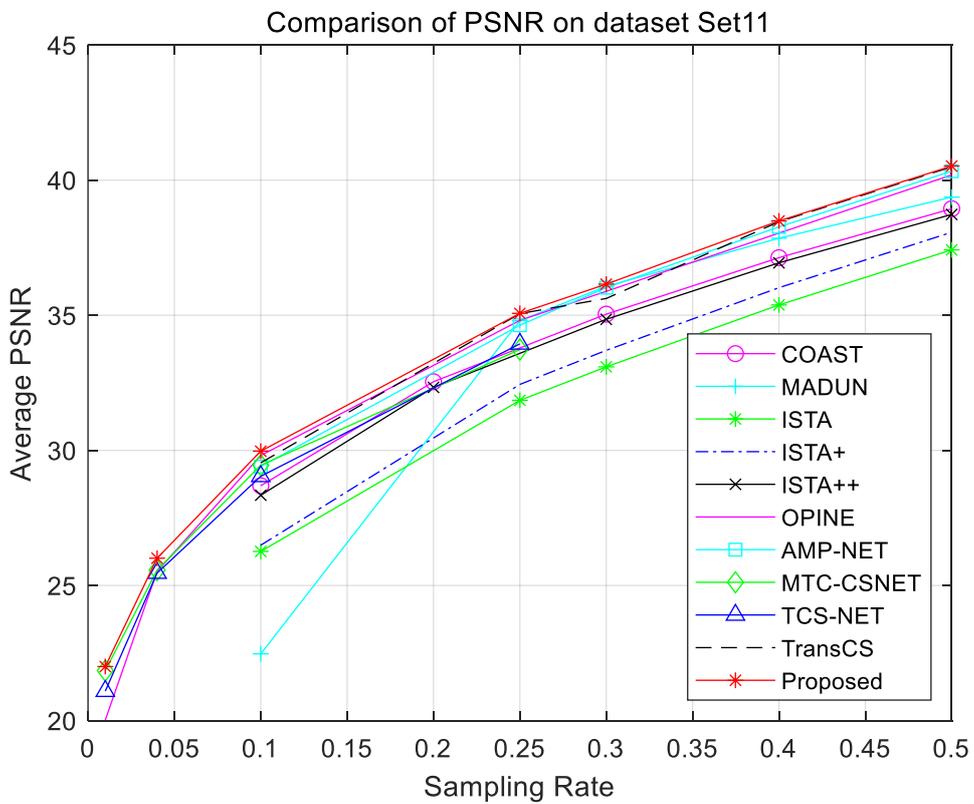

**Fig. 3.** Comparison of PSNR on dataset Set11.

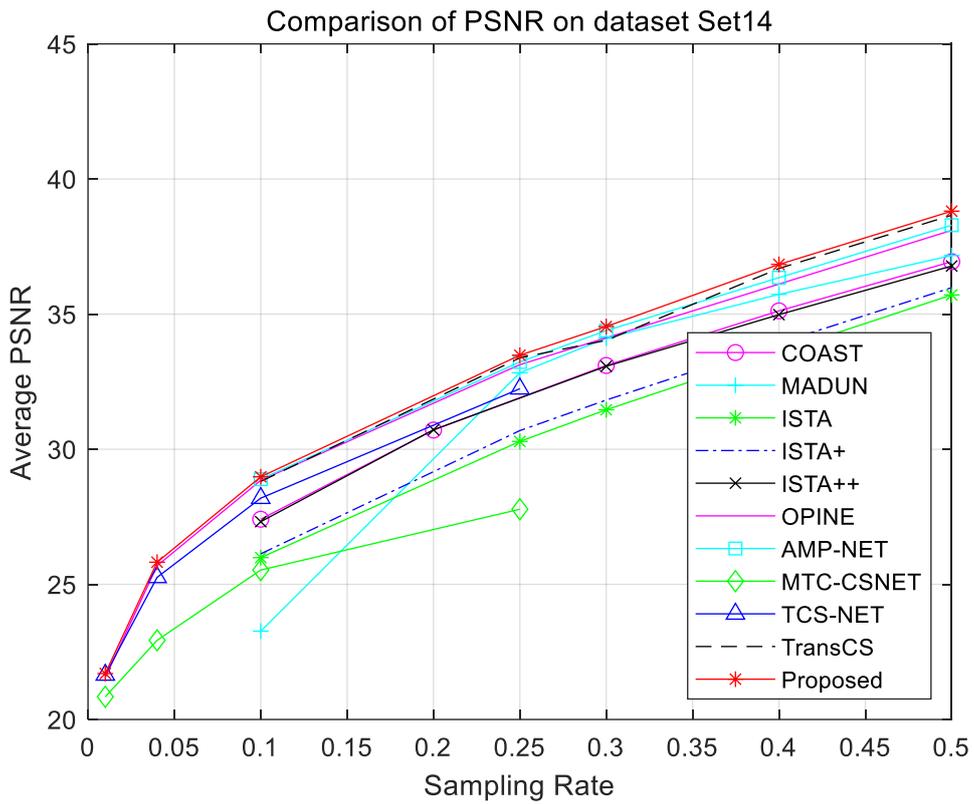

**Fig. 4.** Comparison of PSNR on dataset Set14.

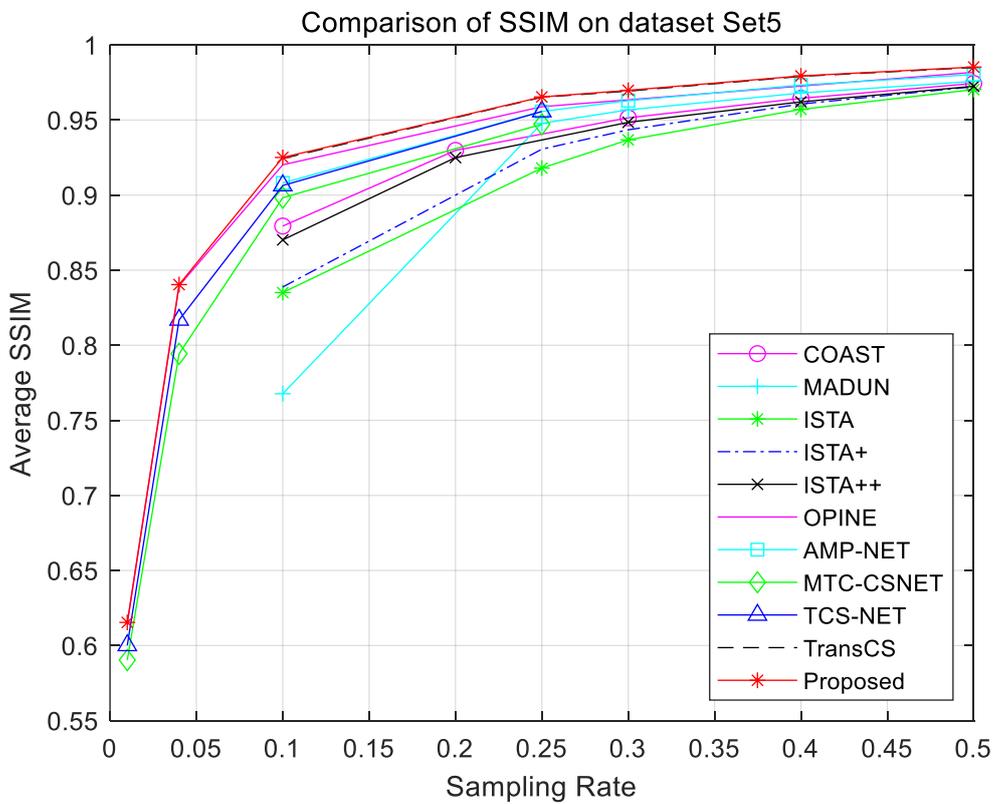

**Fig. 5.** Comparison of SSIM on dataset Set5.

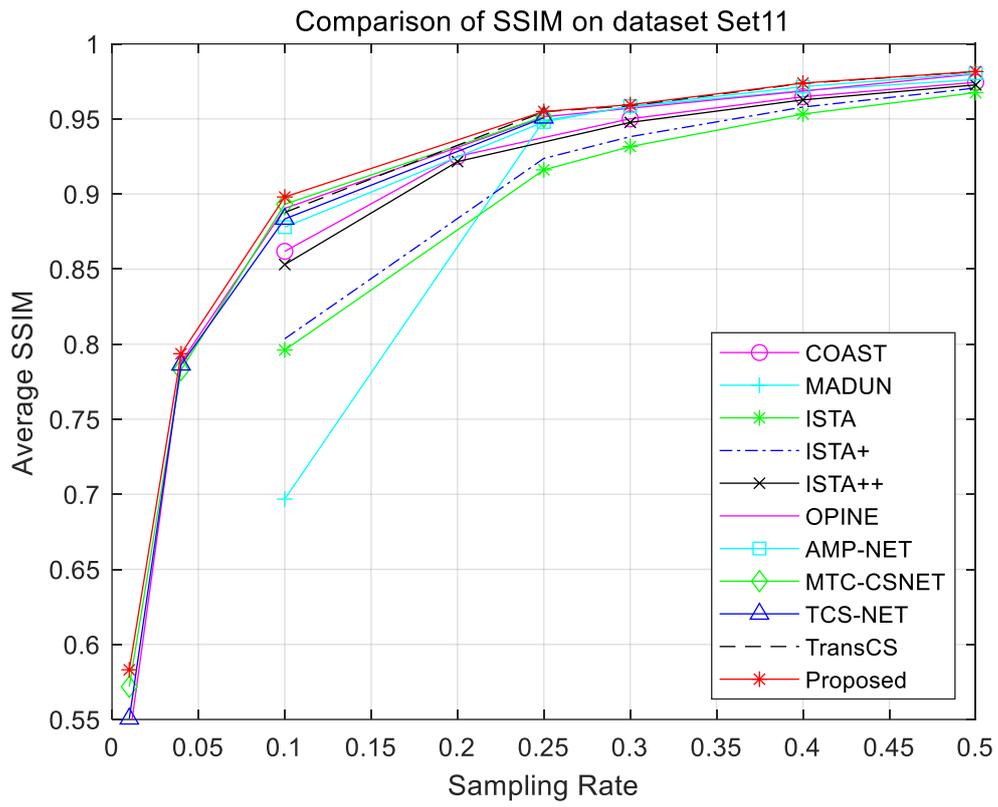

**Fig. 6. Comparison of SSIM on dataset Set11.**

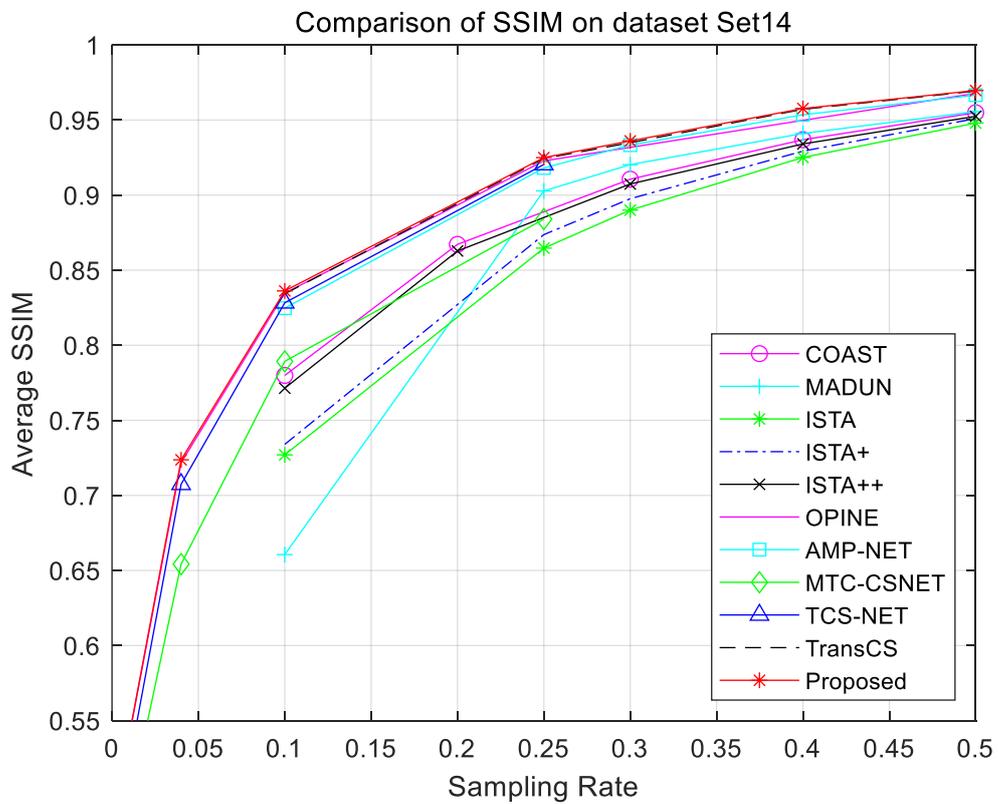

**Fig. 7. Comparison of SSIM on dataset Set14.**

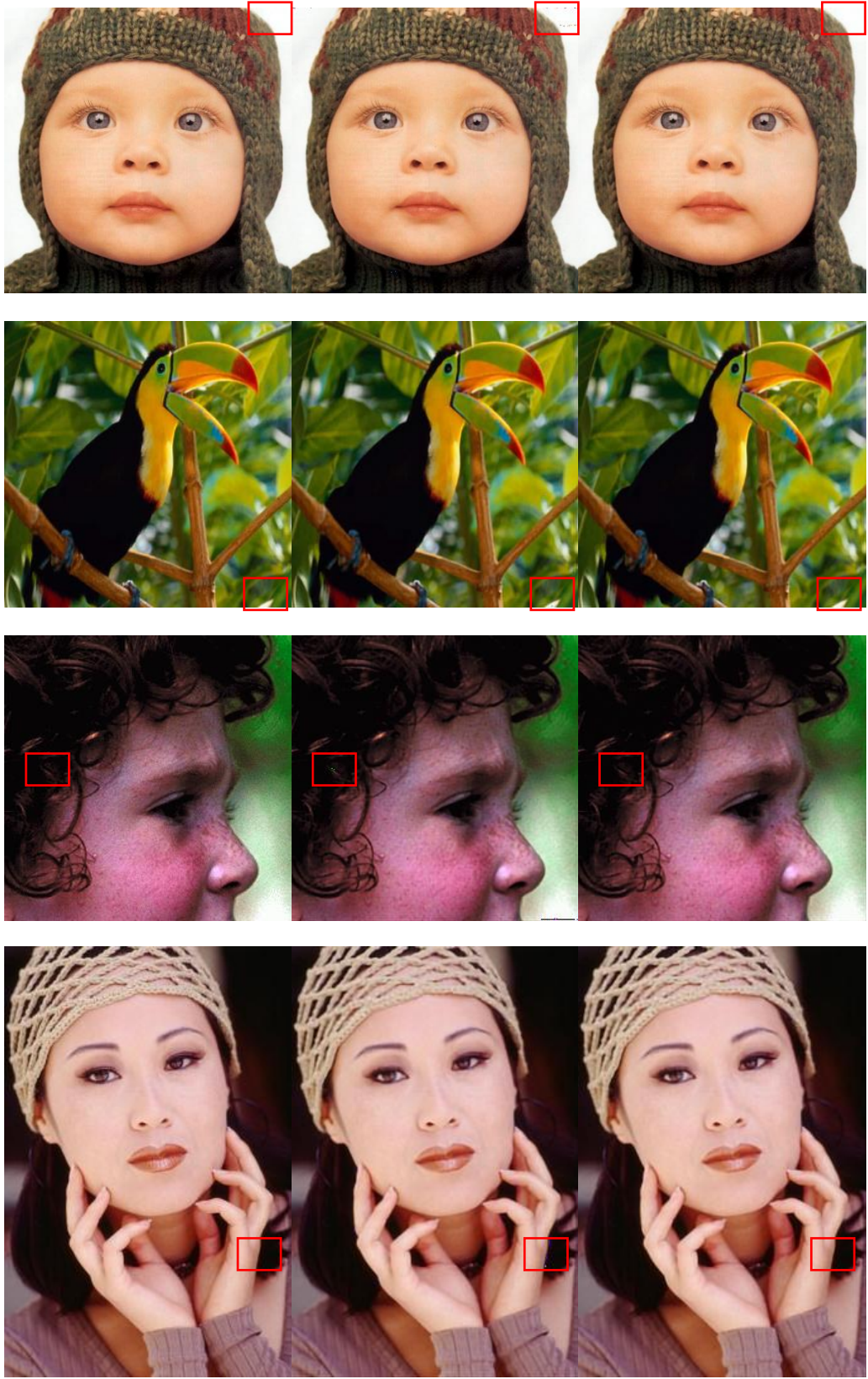

Fig. 8. Comparison with MTC-CSNET on dataset Set5 at the sampling rate of 0.25 [56].

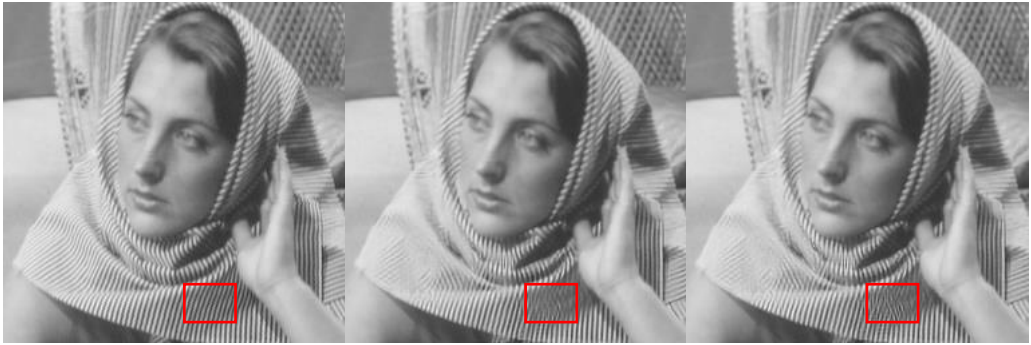
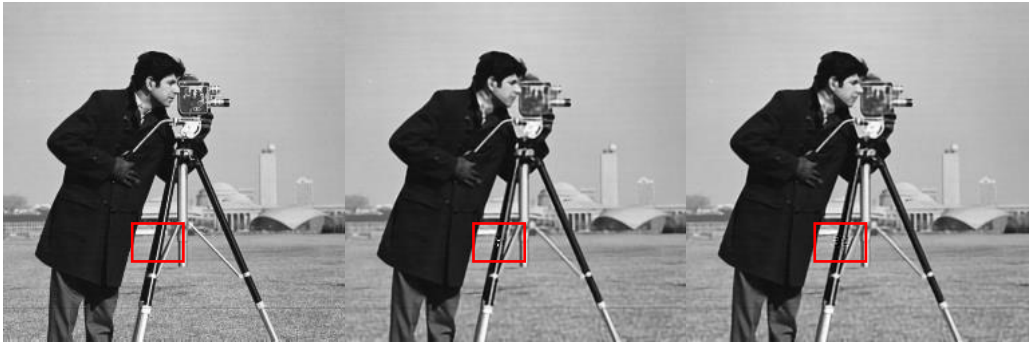
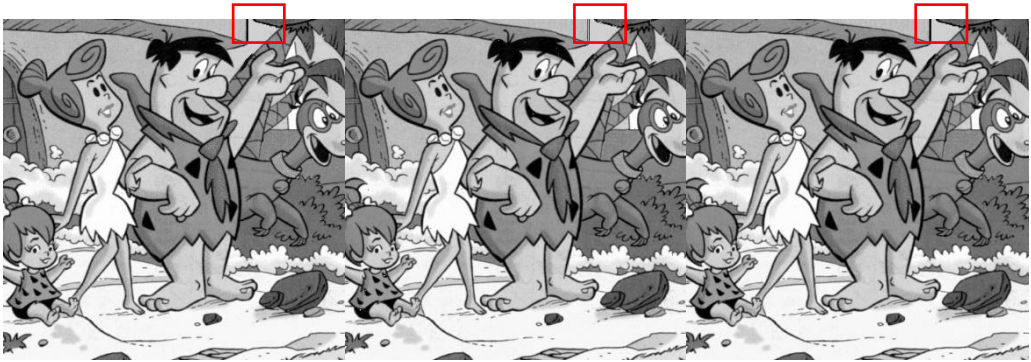
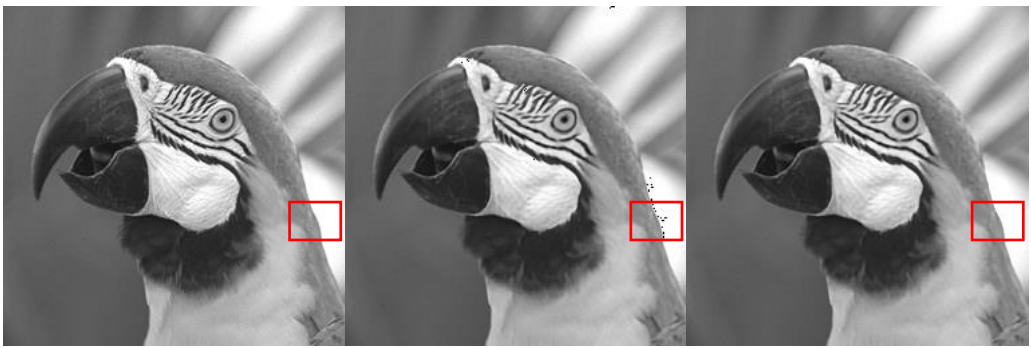
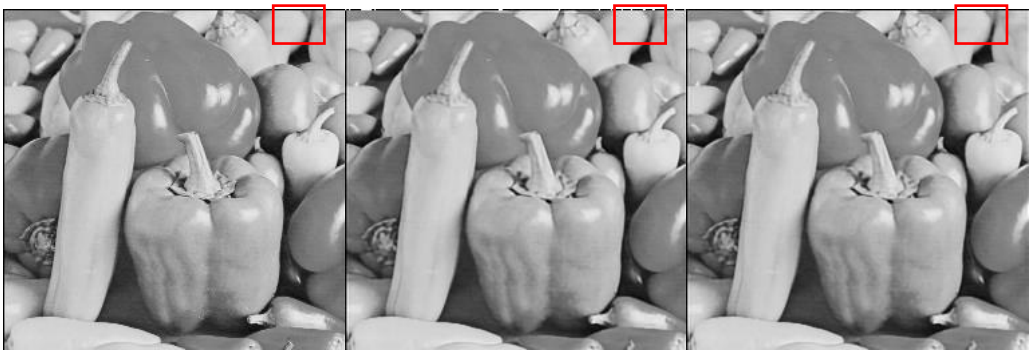

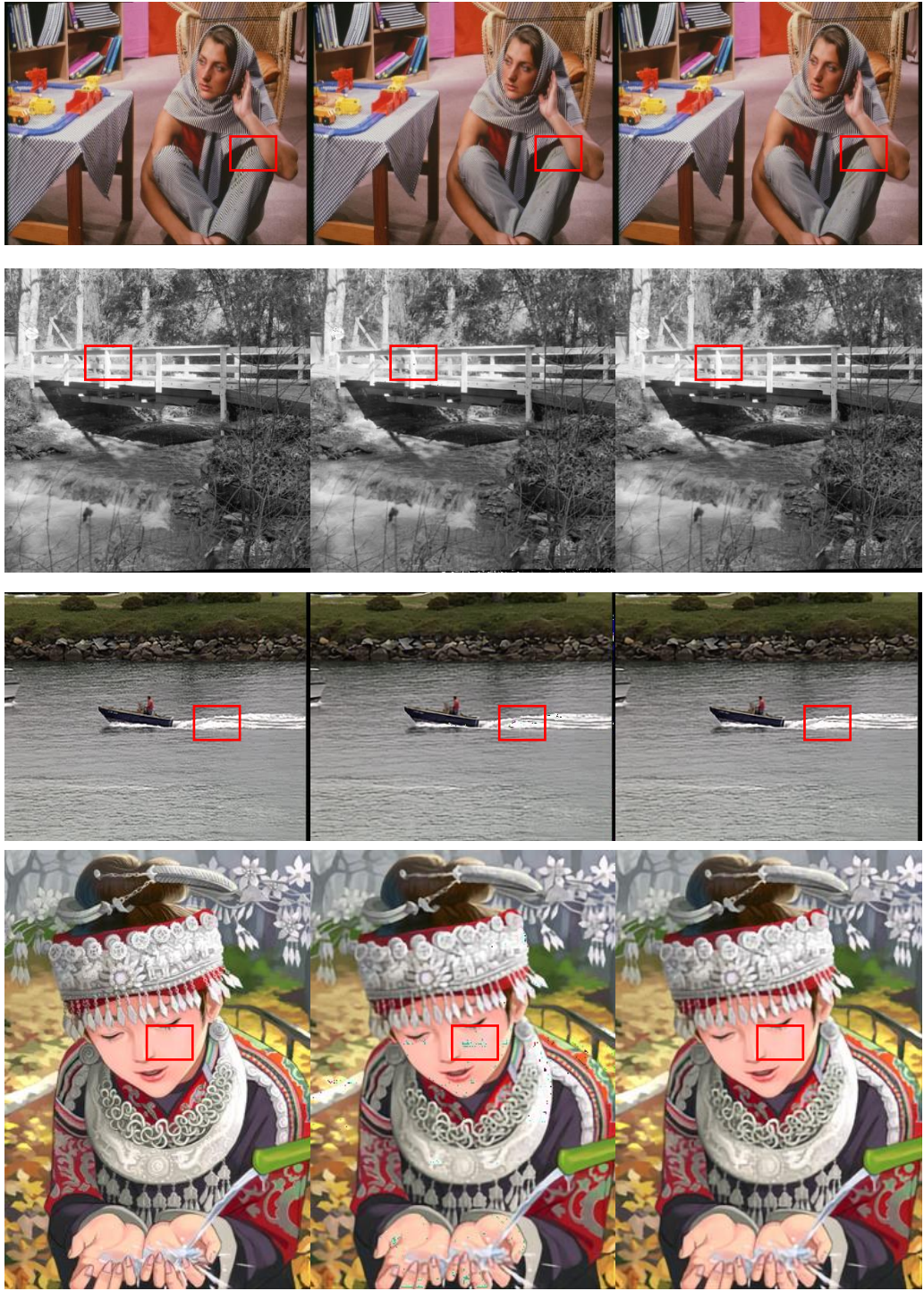

Fig. 9. Comparison with MTC-CSNET on dataset Set11 at the sampling rate of 0.25 [56].

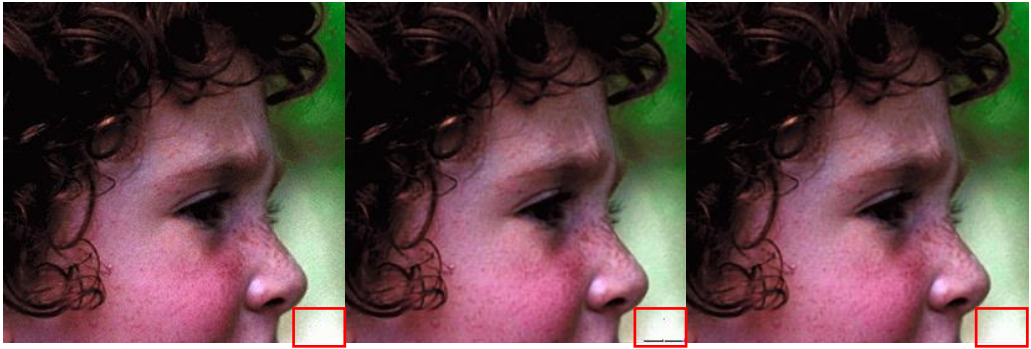
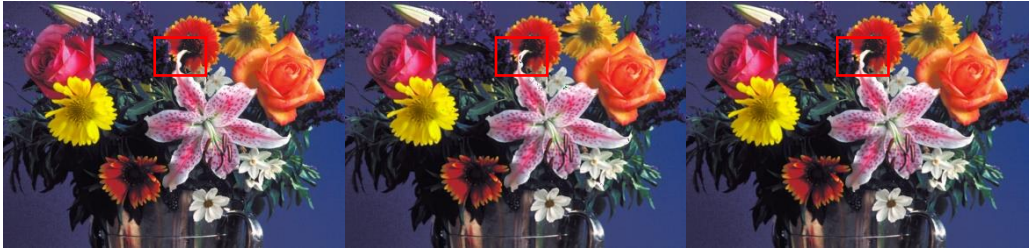
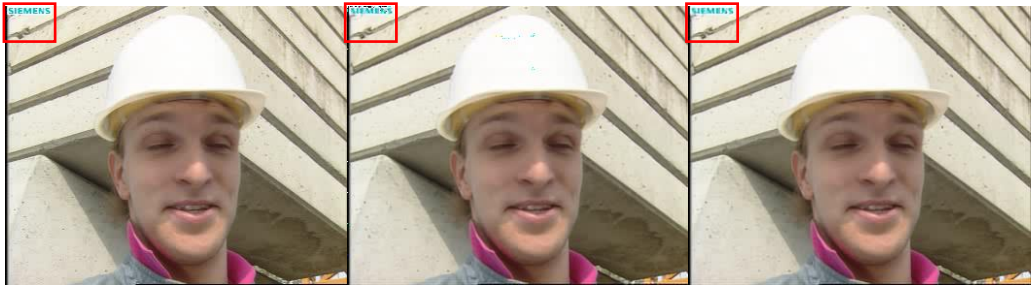
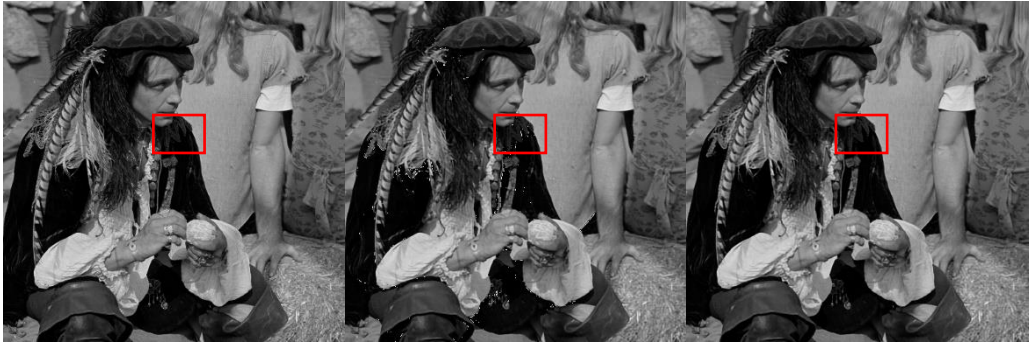
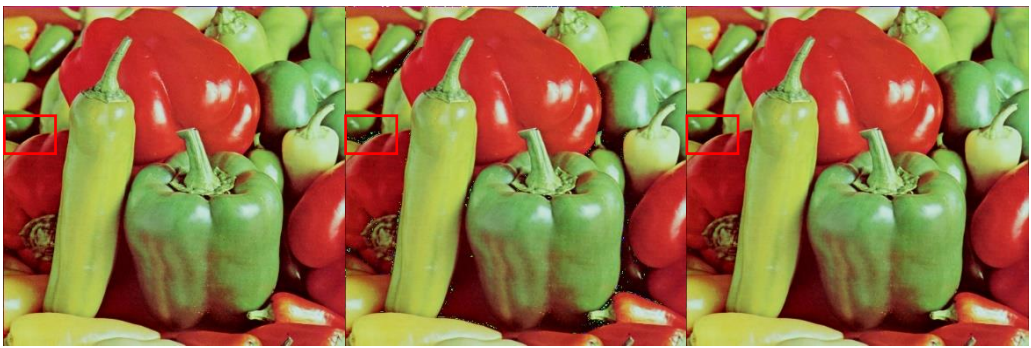

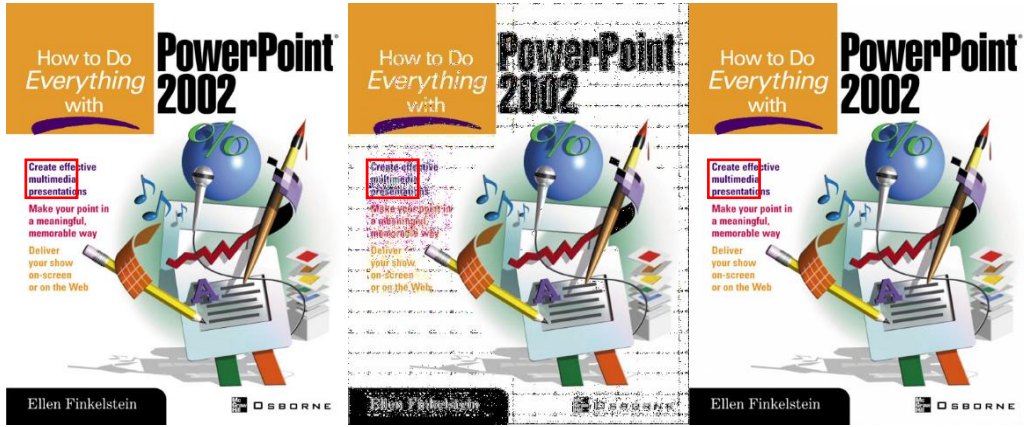

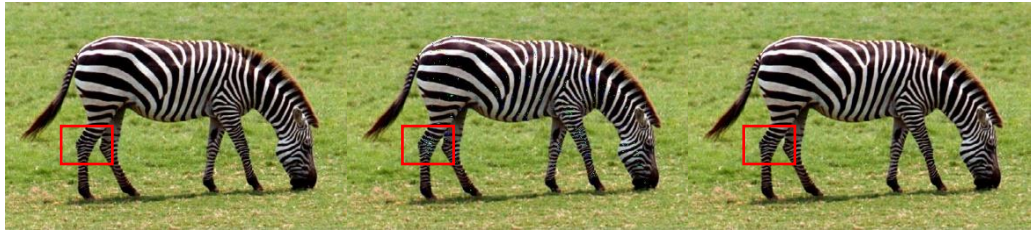

**Fig. 10. Comparison with MTC-CSNET on dataset Set14 at the sampling rate of 0.25 [56].**

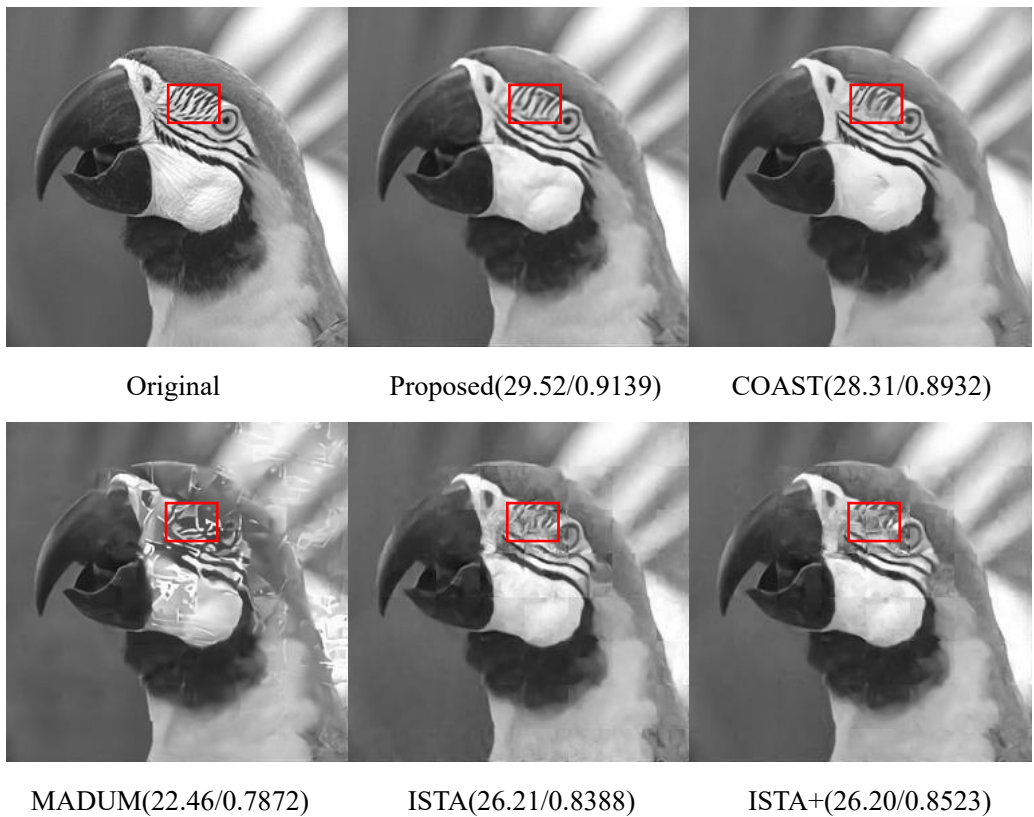

Original     Proposed(29.52/0.9139)     COAST(28.31/0.8932)

MADUM(22.46/0.7872)     ISTA(26.21/0.8388)     ISTA+(26.20/0.8523)

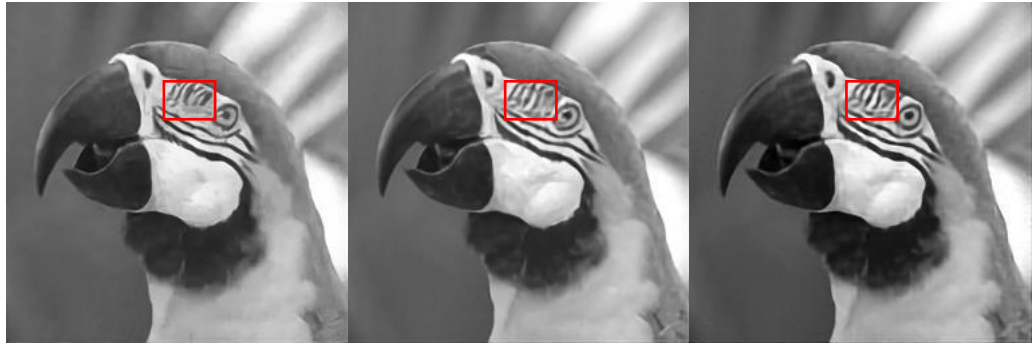

ISTA++(28.09/0.8864)　　OPINE(29.34/0.9155)　　AMP-NET(29.19/0.9054)

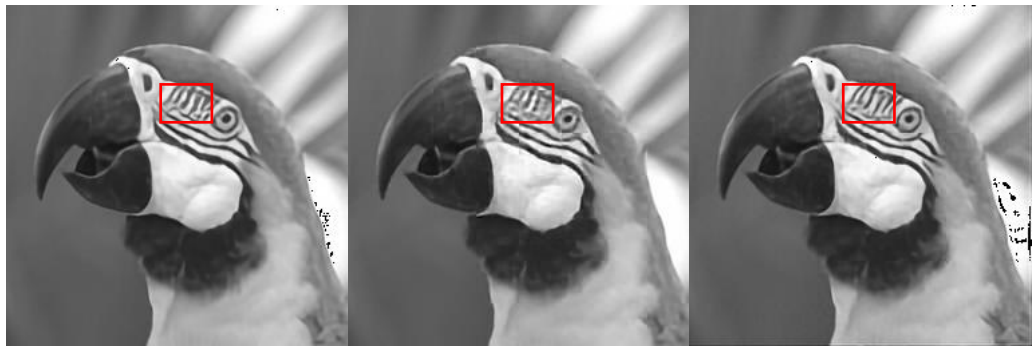

MTC-CSNET(26.66/0.9087)　　TCS-NET(28.57/0.9118)　　TransCS(29.43/0.9134)

**Fig. 11. Comparison with competition approaches on dataset Set11 (Parrots) with the sampling rate of 0.10.**

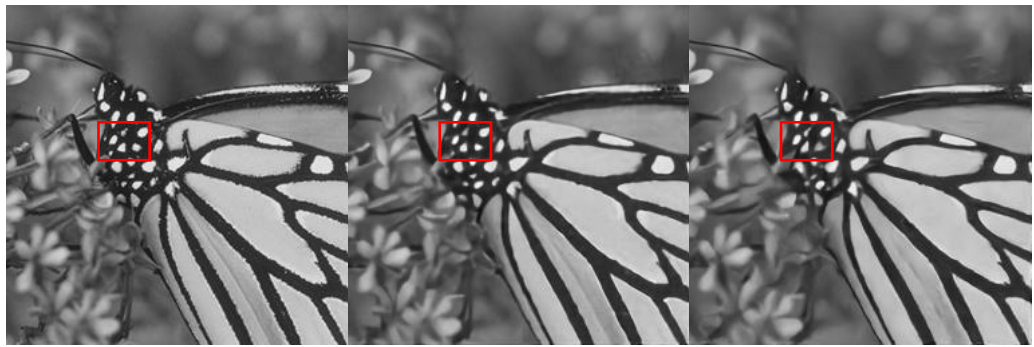

Original　　Proposed(30.09/0.9352)　　COAST(27.95/0.8993)

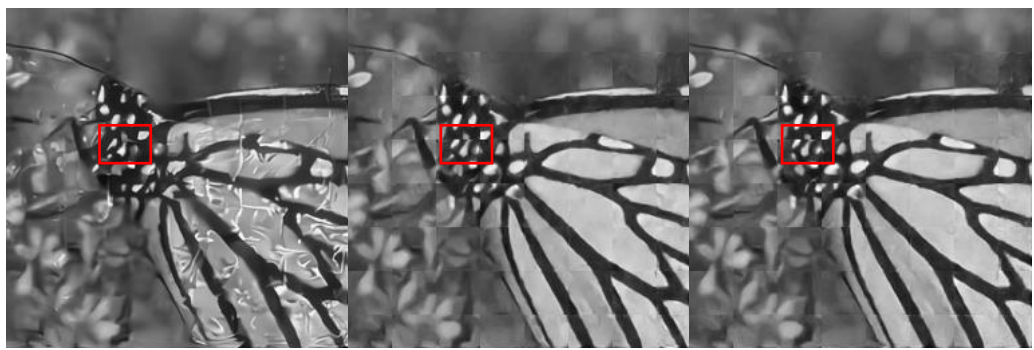

MADUM(21.18/0.7473)　　ISTA(25.58/0.8304)　　ISTA+(25.72/0.8366)

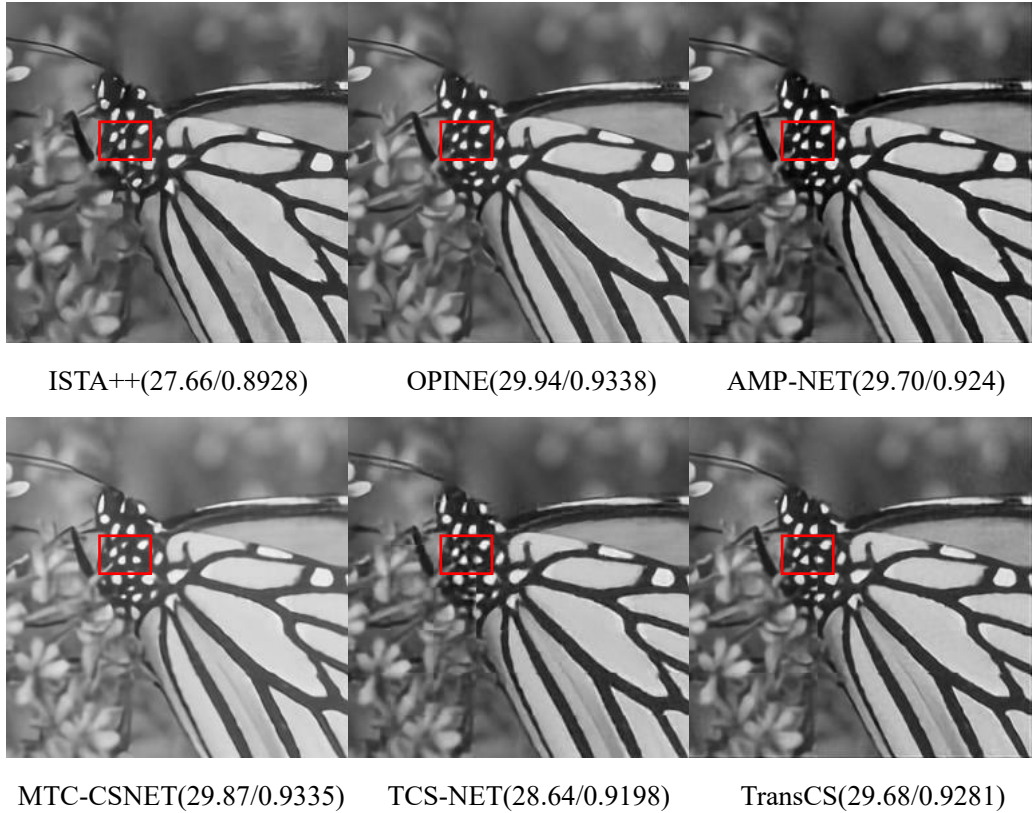

| ISTA++(27.66/0.8928) | OPINE(29.94/0.9338) | AMP-NET(29.70/0.924) |
| MTC-CSNET(29.87/0.9335) | TCS-NET(28.64/0.9198) | TransCS(29.68/0.9281) |

**Fig. 12. Comparison with competition approaches on dataset Set11 (Monarch) with the sampling rate of 0.10.**

## 5 Conclusions

This paper improves the reconstruction performance of conventional image compressive sensing algorithms by incorporating a closed-loop negative feedback mechanism. The closed-loop structure consists of measurement unit, recovery unit, summator, multiplier and delayer. The theory analysis of negative feedback system is carried out. The approximate mathematical proof the effectiveness of the proposed method is offered. The simulation experiments are conducted on more than 3 image datasets and 10 competition approaches. The experimental results show that the proposed method is superior to the competition approaches in recovery quality and can efficiently overcome recovery errors brought by the competition approaches.

In our future work, currently unavailable lossless image compressive sensing will be explored.

## References

[1] Yang Y, Sun J, Li HB, Xu ZB; DMM-CSNet: A Deep Learning Approach for Image Compressive Sensing; IEEE Transactions on Pattern Analysis and Machine Intelligence, 2020, 42(3) , 521-538.
[2] Shi WZ, Jiang F, Liu SH, Teramoto A, Zhao DB; Image Compressed Sensing Using Convolutional Neural Network; IEEE Transactions on Image Processing, 2020, 29, 375-388.
[3] Tavares CA, Santos TMR, Lemes NHT, dos Santos JPC, Ferreira JC, Braga JP; Solving Ill-Posed Problems Faster Using Fractional-Order Hopfield Neural Network; Journal of Computational and Applied Mathematics, 2021, 381, 112984, 1-14.
[4] Zhang Y, Hofmann B; On the Second-Order Asymptotical Regularization of Linear Ill-Posed

Inverse Problems; Applicable Analysis, 2020, 99(6), 1000-1025.

[5] Adler J, Oktem O; Solving Ill-Posed Inverse Problems Using Iterative Deep Neural Networks; Inverse Problems, 2017, 33(12), 1-10.

[6] Huang WK, Zhou FB, Zou T, Lu PW, Xue YH, Liang JJ, Dong YK, Alternating Positive and Negative Feedback Control Model Based on Catastrophe Theories, Mathematics, 2021, 9, 22, 2878.

[7] Li LX, Fang Y, Liu LW, Peng HP, Kurths J, Yang YX; Overview of Compressed Sensing: Sensing Model, Reconstruction Algorithm, and Its Applications; Applied Sciences-Basel, 2020, 10 (17), 1-19.

[8] Monika R, Samiappan D, Kumar R; Adaptive Block Compressed Sensing - A Technological Analysis and Survey on Challenges, Innovation directions and Applications; Multimedia Tools and Applications, 2021, 80(3), 4751-4768.

[9] Chen QP, Shah NJ, Worthoff WA; Compressed Sensing in Sodium Magnetic Resonance Imaging: Techniques, Applications, and Future Prospects; Journal of Magnetic Resonance Imaging, 2021, Online, 1-10.

[10] Bustin A, Fuin N, Botnar RM, Prieto C; From Compressed-Sensing to Artificial Intelligence-Based Cardiac MRI Reconstruction; Frontiers in Cardiovascular Medicine, 2020, 7, 1-19.

[11] Chul YJ; Compressed Sensing MRI: A Review from Signal Processing Perspective; BMC Biomedical Engineering, 2019, 1(8), 1-17.

[12] Yang JG, Jin T, Xiao C, Huang XT; Compressed Sensing Radar Imaging: Fundamentals, Challenges, and Advances; Sensors, 2019, 19(14), 1-19.

[13] Cao BH, Li SZ, Enze C, Fan MB, Gan FX; Progress in Terahertz Imaging Technology; Spectroscopy and Spectral Analysis, 2020, 40(9), 2686-2695.

[14] Ke J, Zhang LX, Zhou Q; Applications of Compressive Sensing in Optical Imaging; Acta Optica Sinica, 2020, 40(1), 1-10.

[15] Hirsch L, Gonzalez MG, Vega LR, A Comparative Study of Time Domain Compressed Sensing Techniques for Optoacoustic Imaging, IEEE Latin America Transactions, 2022, 20, 6, 1018-1024.

[16] Wang J, Tong ZS, Hu CY, Xu MC, Huang ZF; Some Mathematical Problems in Ghost Imaging; Acta Optica Sinica, 2020, 40(1), 1-10.

[17] Yousufi M, Amir M, Javed U, Tayyib M, Abdullah S, Ullah H, Qureshi IM, Alimgeer KS, Akram MW, Khan KB; Application of Compressive Sensing to Ultrasound Images: A Review; Biomed Research International, 2019, 2019, 1-15.

[18] Xie YR, Castro DC, Rubakhin SS, Sweedler JV, Lam F, Enhancing the Throughput of FT Mass Spectrometry Imaging Using Joint Compressed Sensing and Subspace Modeling, Analytical Chemistry, 2022, 94, 13, 5335-5343.

[19] Oiknine Y, August I, Farber V, Gedalin D, Stern A; Compressive Sensing Hyperspectral Imaging by Spectral Multiplexing with Liquid Crystal; Journal of Imaging, 2019, 5(1), 1-17.

[20] Calisesi G, Ghezzi A, Ancora D, D'Andrea C, Valentini G, Farina A, Bassi A; Compressed Sensing in Fluorescence Microscopy; Progress in Biophysics & Molecular Biology, 2022, 168, 66-80.

[21] Monika R, Dhanalakshmi S, Kumar R, Narayanamoorthi R, Lai KW, An Efficient Adaptive Compressive Sensing Technique for Underwater Image Compression in IoUT, Wireless Networks, 2022.

[22] Edgar MP, Gibson GM, Padgett MJ; Principles and Prospects for Single-Pixel Imaging; Nature Photonics, 2019, 13(1), 13-20.

[23] Xiao XY, Chen LY, Zhang XZ, Wang C, Lan RJ, Ren C, Cao DZ; Review on Single-Pixel Imaging and Its Probability Statistical Analysis; Laser & Optoelectronics Progress, 2021, 58(10), 1-10.
[24] Gibson GM, Johnson SD, Padgett MJ; Single-Pixel Imaging 12 Years on: a Review; Optics Express, 2020, 28(19), 28190-28208.
[25] Zanotto L, Piccoli R, Dong JL, Morandotti R, Razzari L; Single-Pixel Terahertz Imaging: a Review; Opto-Electronic Advances, 2020, 3(9), 1-15.
[26] Liu F, Yao XR, Liu XF, Zhai GJ; Single-Photon Time-Resolved Imaging Spectroscopy Based on Compressed Sensing; Laser & Optoelectronics Progress, 2021, 58(10), 1-10.
[27] Zhang ML, Compressive Sensing Acquisition with Application to Marchenko Imaging, Pure and Applied Geophysics, Early Access, 2022.
[28] Ravishankar S, Ye JC, Fessler JA; Image Reconstruction: From Sparsity to Data-Adaptive Methods and Machine Learning; Proceedings of the IEEE, 2020, 108(1), 86-109.
[29] Xie YT, Li QZ, A Review of Deep Learning Methods for Compressed Sensing Image Reconstruction and Its Medical Applications, Electronics, 2022, 11, 4, 586.
[30] Saideni W, Helbert D, Courreges F, Cances JP, An Overview on Deep Learning Techniques for Video Compressive Sensing, Applied Sciences-BASEL, 2022, 12, 5, 2734.
[31] Khosravy M, Cabral TW, Luiz MM, Gupta N, Crespo RG, Random Acquisition in Compressive Sensing: A Comprehensive Overview, International Journal of Ambient Computing and Intelligence, 2021, 12, 3, 140-165.
[32] Mishra I, Jain S, Soft Computing Based Compressive Sensing Techniques in Signal Processing: A Comprehensive Review, Journal of Intelligent Systems, 2021, 30, 1, 312-326.
[33] Chen YT, Schonlieb CB, Lio P, Leiner T, Dragotti PL, Wang G, Rueckert D, Firmin D, Yang G; AI-Based Reconstruction for Fast MRI-A Systematic Review and Meta-Analysis; Proceedings of the IEEE, 2022, 110(2), 224-245.
[34] Zhang ML, Zhang MY, Zhang F, Chaddad A, Evans A, Robust Brain MR Image Compressive Sensing via Re-Weighted Total Variation and Sparse Regression, Magnetic Resonance Imaging, 2022, 85, 271–286.
[35] Zhang JC, Han LL, Sun JZ, Wang ZK, Xu WL, Chu YH, Xia L, Jiang MF, Compressed Sensing Based Dynamic MR Image Reconstruction by Using 3D-Total Generalized Variation and Tensor Decomposition: k-t TGV-TD, BMC Medical Imaging, 2022, 22, 1, 101.
[36] Yin Z, Shi WZ, Wu ZC, Zhang J, Multilevel Wavelet-Based Hierarchical Networks for Image Compressed Sensing, Pattern Recognition, 2022, 129, 108758.
[37] Yin Z, Wu ZC, Zhang J, A Deep Network Based on Wavelet Transform for Image Compressed Sensing, Circuits Systems and Signal Processing, 2022.
[38] Lv MJ, Ma L, Ma JC, Chen WF, Yang J, Ma XY, Cheng Q, Fast, Super-Resolution Sparse Inverse Synthetic Aperture Radar Imaging via Continuous Compressive Sensing, IET Signal Processing, 2022, 16, 3, 310-326.
[39] Sun M, Tao JX, Ye ZF, Qiu BS, Xu JZ, Xi CF; An Algorithm Combining Analysis-based Blind Compressed Sensing and Nonlocal Low-rank Constraints for MRI Reconstruction; Current Medical Imaging Reviews, 2019, 15(3), 281-291.
[40] Li HG; Compressive Domain Spatial-Temporal Difference Saliency-Based Realtime Adaptive Measurement Method for Video Recovery; IET Image Processing, 2019, 13(11), 2008-2017.
[41] Suantai S, Noor MA, Kankam K, Cholamjiak P; Novel Forward-Backward Algorithms for

Optimization and Applications to Compressive Sensing and Image Inpainting; Advances in Difference Equations, 2021, 2021(1), 1-22.

[42] Mardani M, Gong EH, Cheng JY, Vasanawala SS, Zaharchuk G, Xing L, Pauly JM; Deep G5nerative Adversarial Neural Networks for Compressive Sensing MRI; IEEE Transactions on Medical Imaging, 2019, 38(1), 167-179.

[43] Li WZ, Zhu AH, Xu YG, Yin HS, Hua G, A Fast Multi-Scale Generative Adversarial Network for Image Compressed Sensing, Entropy, 2022, 24, 6.

[44] Zeng GS, Guo Y, Zhan JY, Wang Z, Lai ZY, Du XF, Qu XB, Guo D; A Review on Deep Learning MRI Reconstruction without Fully Sampled k-Space; BMC Medical Imaging, 2021, 21(1), 1-11.

[45] Han Y, Sunwoo L, Ye JC; k-Space Deep Learning for Accelerated MRI; IEEE Transactions on Medical Imaging, 2020, 39(2), 377-386.

[46] Kravets V, Stern A, Progressive Compressive Sensing of Large Images with Multiscale Deep Learning Reconstruction, Scientific Reports, 2022, 12, 1, 7228.

[47] Wang ZB, Qin YL, Zheng H, Wang RF, Multiscale Deep Network for Compressive Sensing Image Reconstruction, Journal of Electronic Imaging, 2022, 31, 1, 013025.

[48] Gan HP, Gao Y, Liu CY, Chen HW, Zhang T, Liu F, AutoBCS: Block-Based Image Compressive Sensing With Data-Driven Acquisition and Noniterative Reconstruction, IEEE Transactions on Cybernetics, 2022, Early Access.

[49] You D, Zhang J, Xie JF, Chen B, Ma SW, COAST: COntrollable Arbitrary-Sampling NeTwork for Compressive Sensing, IEEE Transactions on Image Processing, 2021, 30, 6066-6080.

[50] Song JC, Chen B, Zhang J, Memory-Augmented Deep Unfolding Network for Compressive Sensing, 29th ACM International Conference on Multimedia (ACM MM), 2021.

[51] Zhang J, Ghanem B, ISTA-Net: Interpretable Optimization-Inspired Deep Network for Image Compressive Sensing, CVPR, 2018.

[52] You D, Xie JF, Zhang J, ISTA-Net++: Flexible Deep Unfolding Network for Compressive Sensing, IEEE International Conference on Multimedia and Expo (ICME), 2021, 1–6.

[53] Zhang J, Zhao C, Gao W, Optimization-Inspired Compact Deep Compressive Sensing, IEEE Journal of Selected Topics in Signal Processing (JSTSP) , 2020, 14, 4, 765-774.

[54] Zhang J, Zhang ZY, Xie JF, Zhang YB, High-Throughput Deep Unfolding Network for Compressive Sensing MRI, IEEE Journal of Selected Topics in Signal Processing, 2022, Early Access.

[55] Zhang ZH, Liu YP, Liu JN, Wen F, Zhu C, AMP-Net: Denoising-Based Deep Unfolding for Compressive Image Sensing, IEEE Transactions on Image Processing, 2021,30,1487-1500.

[56] shen65536@mail.nwpu.edu.cn, MTC-CSNet: Marrying Transformer and Convolution for Image Compressed Sensing, https://github.com/EchoSPLab/MTC-CSNet, 2022.

[57] shen65536@mail.nwpu.edu.cn, TCS-Net: From Patch to Pixel: A Transformer-based Hierarchical Framework for Compressive Image Sensing, https://github.com/CompressiveLab/ TCS-Net, 2022.

[58] shen65536@mail.nwpu.edu.cn, TransCS: A Transformer-based Hybrid Architecture for Image Compressed Sensing, https://github.com/myheuf/TransCS, 2022.

[59] Harada Y, Kanemoto D, Inoue T, Maida O, Hirose T, Image Quality Improvement for Capsule Endoscopy Based on Compressed Sensing with K-SVD Dictionary Learning, IEICE Transactions on Fundamentals of Electronics Communications and Computer Sciences, 2022, E105A, 4, 743-

747.

[60] Ueki W, Nishii T, Umehara K, Ota J, Higuchi S, Ohta Y, Nagai Y, Murakawa K, Ishida T, Fukuda T, Generative Adversarial Network-Based Post-Processed Image Super-Resolution Technology for Accelerating Brain MRI: Comparison with Compressed Sensing, ACTA RADIOLOGICA, 2022, 02841851221076330.

[61] Fang CJ, Chen JY, Chen SL, Image Denoising Algorithm of Compressed Sensing Based on Alternating Direction Method of Multipliers, International Journal of Modeling Simulation and Scientific Computing, 2022, 13, 01, 2250009.

[62] El MA, Ouahabi A, Moulay MS, Image Denoising Using a Compressive Sensing Approach Based on Regularization Constraints, Sensors, 2022, 22, 6, 2199.

[63] Pham CDK, Yang J, Zhou JJ, CSIE-M: Compressive Sensing Image Enhancement Using Multiple Reconstructed Signals for Internet of Things Surveillance Systems, IEEE Transactions on Industrial Informatics, 2022, 18, 2, 1271-1281.

[64] Zhang Y., Mao X., Wang J., Liu W., DEMO: A Flexible Deartifacting Module for Compressed Sensing MRI, IEEE Journal of Selected Topics in Signal Processing, 2022, 16, 4, 725-736.

## Acknowledgements

The authors would very much like to thank all the authors of the 10 competition approaches for selflessly releasing their source codes of image compressive sensing on GitHub website. We can easily implement our method based on the open source codes.

## Declaration

The authors declare that they have no known competing financial interests or personal relationships that could have appeared to influence the work reported in this paper.

## Biographies

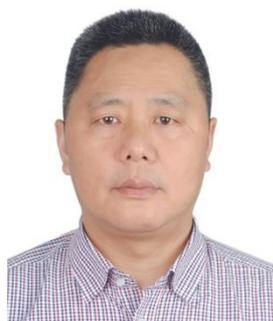

**Honggui Li** received a B.S. degree in electronic science and technology from Yangzhou University, and received a Ph.D. degree in mechatronic engineering from Nanjing University of Science and Technology. He is a senior member of the Chinese Institute of Electronics. He is a visiting scholar and a post-doctoral fellow in Institut Supérieur d'Électronique de Paris for one year. He is an associate professor of electronic science and technology and a postgraduate supervisor of electronic science and technology at Yangzhou University. He is a reviewer for some international journals and conferences. He is the author of over 30 refereed journal and conference articles. His current research interests include machine learning, deep learning, computer vision, and embedded computing.

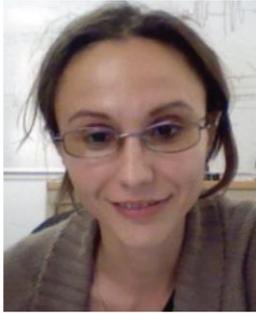

**Maria Trocan** received her M.Eng. in Electrical Engineering and Computer Science from Politehnica University of Bucharest, the Ph.D. in Signal and Image Processing from Telecom ParisTech and the Habilitation to Lead Researches (HDR) from Pierre & Marie Curie University (Paris 6). She has joined Joost - Netherlands, where she worked as research engineer involved in the design and development of video transcoding systems. She is firstly Associate Professor, then Professor at Institut Superieur d'Electronique de Paris (ISEP). She is Associate Editor for Springer Journal on Signal, Image and Video Processing and Guest Editor for several journals (Analog Integrated Circuits and Signal Processing, IEEE Communications Magazine etc.). She is an active member of IEEE France and served as counselor for the ISEP IEEE Student Branch, IEEE France Vice-President responsible of Student Activities and IEEE Circuits and Systems Board of Governors member, as Young Professionals representative. Her current research interests focus on image and video analysis & compression, sparse signal representations, machine learning and fuzzy inference.

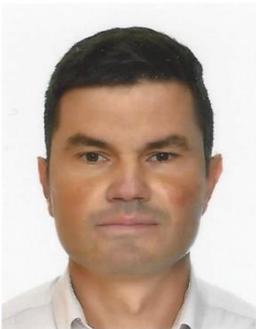

**Dimitri Galayko** received the bachelor's degree from Odessa State Polytechnic University in Ukraine, the master's degree from the Institute of Applied Sciences of Lyon in France, and the Ph.D. degree from University Lille in France. He made his Ph.D. thesis in the Institute of Microelectronics and Nanotechnologies. His Ph.D. dissertation was on the design of micro-electromechanical silicon filters and resonators for radio-communications. He is an Associate Professor with the LIP6 research laboratory of Sorbonne University in France. His research interests include study, modelling, and design of nonlinear integrated circuits for sensor interface and for mixed-signal applications. His research interests also include machine learning and fuzzy computing.

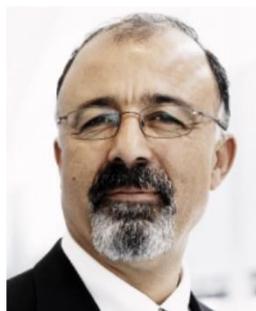

**Mohamad Sawan** (Fellow, IEEE) received the Ph.D. degree in electrical engineering from the University of Sherbrooke, Sherbrooke, QC, Canada, in 1990. He was a Chair Professor awarded with the Canada Research Chair in Smart Medical Devices (2001–2015), and was leading the Microsystems Strategic Alliance of Quebec - ReSMiQ (1999–2018). He is a Professor of Microelectronics and Biomedical Engineering, in leave of absence from Polytechnique Montréal, Canada. He joined Westlake University, Hangzhou, China, in January 2019, where he is a Chair Professor, Founder, and the Director of the Center for Biomedical Research And INnovation (CenBRAIN). He has published more than 800 peer-reviewed articles, two books, ten book chapters, and 12 patents. He founded and chaired the IEEE-Solid State Circuits Society Montreal Chapter (1999–2018) and founded the Polystim Neurotech Laboratory, Polytechnique Montréal (1994–present), including two major research infrastructures intended to build advanced Medical devices. He is the Founder of the International IEEE-NEWCAS Conference, and the Co-Founder of the International IEEE-BioCAS, ICECS, and LSC conferences. He is a Fellow of the Canadian Academy of Engineering, and a Fellow of the Engineering Institutes of Canada. He is also the

"Officer" of the National Order of Quebec. He has served as a member of the Board of Governors (2014–2018). He is the Vice-President Publications (2019–present) of the IEEE CAS Society. He received several awards, among them the Queen Elizabeth II Golden Jubilee Medal, the Barbara Turnbull 2003 Award for spinal-cord research, the Bombardier and Jacques-Rousseau Awards for academic achievements, the Shanghai International Collaboration Award, and the medal of merit from the President of Lebanon for his outstanding contributions. He was hosted in Montreal as the General Chair, the 2016 IEEE International Symposium on Circuits and Systems (ISCAS), and hosted as the General Chair of the 2020 IEEE International Medicine, Biology and Engineering Conference (EMBC). Dr. Sawan was the Deputy Editor-in-Chief of the IEEE TRANSACTIONS ON CIRCUITS AND SYSTEMS-II: EXPRESS BRIEFS (2010–2013); the Co-Founder, an Associate Editor, and the Editor-in-Chief of the IEEE TRANSACTIONS ON BIOMEDICAL CIRCUITS AND SYSTEMS; an Associate Editor of the IEEE TRANSACTIONS ON BIOMEDICALS ENGINEERING; and the International Journal of Circuit Theory and Applications.